\documentclass[sigconf,natbib=true]{acmart}
\AtBeginDocument{%
  \providecommand\BibTeX{{%
    \normalfont B\kern-0.5em{\scshape i\kern-0.25em b}\kern-0.8em\TeX}}}

\copyrightyear{2023} 
\acmYear{2023} 
\setcopyright{rightsretained} 
\acmConference[KDD '23]{Proceedings of the 29th ACM SIGKDD Conference on Knowledge Discovery and Data Mining}{August 6--10, 2023}{Long Beach, CA, USA}
\acmBooktitle{Proceedings of the 29th ACM SIGKDD Conference on Knowledge Discovery and Data Mining (KDD '23), August 6--10, 2023, Long Beach, CA, USA}
\acmDOI{10.1145/3580305.3599370}
\acmISBN{979-8-4007-0103-0/23/08}

\usepackage{enumitem}
\usepackage[ruled]{algorithm2e}
\usepackage{subfigure}
\newcommand{\change}{}

\makeatletter
\gdef\@copyrightpermission{
  \begin{minipage}{0.3\columnwidth}
   \href{https://creativecommons.org/licenses/by-nc-sa/4.0/}{\includegraphics[width=0.90\textwidth]{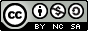}}
  \end{minipage}\hfill
  \begin{minipage}{0.7\columnwidth}
   \href{https://creativecommons.org/licenses/by-nc-sa/4.0/}{This work is licensed under a Creative Commons Attribution-NonCommercial-ShareAlike International 4.0 License.}
  \end{minipage}
}
\makeatother

\begin{document}

\title{Graph Contrastive Learning with Generative Adversarial Network}

\author{Cheng Wu}
\orcid{0009-0002-4481-405X}
\email{wuc22@mails.tsinghua.edu.cn}
\affiliation{%
  \institution{Tsinghua University}
  \city{Beijing}
  \country{China}
}

\author{Chaokun Wang}
\orcid{0000-0002-2986-2574}
\authornote{Chaokun Wang is the corresponding author.}
\email{chaokun@tsinghua.edu.cn}
\affiliation{%
  \institution{Tsinghua University}
  \city{Beijing}
  \country{China}
}

\author{Jingcao Xu}
\email{xjc20@mails.tsinghua.edu.cn}
\orcid{0009-0000-3715-6445}
\affiliation{%
  \institution{Tsinghua University}
  \city{Beijing}
  \country{China}
}

\author{Ziyang Liu}
\email{liuzy21@mails.tsinghua.edu.cn}
\orcid{0009-0007-4238-1533}
\affiliation{%
  \institution{Tsinghua University}
  \city{Beijing}
  \country{China}
}

\author{Kai Zheng}
\orcid{0009-0006-3822-2815}
\email{zhengkai@kuaishou.com}
\affiliation{%
  \institution{Kuaishou Inc.}
  \city{Beijing}
  \country{China}
}

\author{Xiaowei Wang}
\orcid{0009-0003-1112-1027}
\email{wangxiaowei03@kuaishou.com}
\affiliation{%
  \institution{Kuaishou Inc.}
  \city{Beijing}
  \country{China}
}

\author{Yang Song}
\orcid{0000-0002-1714-5527}
\email{yangsong@kuaishou.com}
\affiliation{%
  \institution{Kuaishou Inc.}
  \city{Beijing}
  \country{China}
}

\author{Kun Gai}
\orcid{0000-0002-3636-3618}
\email{gai.kun@qq.com}
\affiliation{%
  \institution{Unaffiliated}
  \city{Beijing}
  \country{China}
}

\renewcommand{\shortauthors}{Cheng Wu et al.}

\begin{abstract}
Graph Neural Networks (GNNs) have demonstrated promising results on exploiting node representations for many downstream tasks through supervised end-to-end training.
To deal with the widespread label scarcity issue in real-world applications, Graph Contrastive Learning (GCL) is leveraged to train GNNs with limited or even no labels by maximizing the mutual information between nodes in its augmented views generated  from the original graph.
However, 
the distribution of graphs remains unconsidered in view generation, resulting in the ignorance of \textit{unseen} edges in most existing literature, which is empirically shown to be able to improve GCL's performance in our experiments.
To this end, we propose to incorporate graph generative adversarial networks (GANs) to learn the distribution of views for GCL, in order to 
i) automatically capture the characteristic of graphs for augmentations, and ii) jointly train the graph GAN model and the GCL model.
Specifically, we present GACN, a novel \textbf{G}enerative \textbf{A}dversarial \textbf{C}ontrastive learning \textbf{N}etwork for graph representation learning.
GACN develops a view generator and a view discriminator to generate augmented views automatically in an adversarial style.
Then, GACN leverages these views to train a GNN encoder with two carefully designed self-supervised learning losses\change{, including the graph contrastive loss and the Bayesian personalized ranking Loss.}
Furthermore, we design an optimization framework to train all GACN modules jointly.
Extensive experiments on seven real-world datasets show that GACN is able to generate high-quality augmented views for GCL and is superior to twelve state-of-the-art baseline methods.
Noticeably, our proposed GACN surprisingly discovers that the generated views in data augmentation finally conform to the well-known \textit{preferential attachment} rule in online networks.
\end{abstract}





\maketitle

\section{Introduction}
In recent years, graph representation learning has attracted increasing attention from both academia and industry to deal with network-based data~\citep{perozzi2014deepwalk, tang2015line, node2vec-kdd2016, gu2022hybridgnn, wu2023SUPA}.
Graph Neural Networks (GNNs)~\citep{kipf2016semi, hamilton2017inductive, he2020lightgcn} have shown effectiveness in supervised end-to-end training.
However, task-specific labels can be extremely scarce for graph datasets~\citep{zitnik2018prioritizing,hu2019strategies}.
To this end, research efforts start exploring self-supervised learning for GNNs, where only limited or even no labels are needed~\citep{wu2021self}.


Recently, graph contrastive learning (GCL)~\citep{velickovic2019deep,you2020graph,zhu2020deep,wu2021self} has become one of the most popular self-supervised approaches, which leverages the mutual information maximization principle (InfoMax)~\citep{linsker1988self} to maximize the correspondence between the representations of a graph (or a node) in its different augmented views.
There are a large amount of view augmentation strategies explored by different GCL methods, including node dropping, edge perturbation, subgraph sampling and feature masking.
Furthermore, views can be generated by random sampling~\citep{you2020graph}, or under the guide of domain knowledge~\cite{hassani2020contrastive,zhu2021graph}, or by a view learner~\cite{suresh2021adversarial}.

\begin{figure}[ht]
\centering
\includegraphics[width=1.0\linewidth]{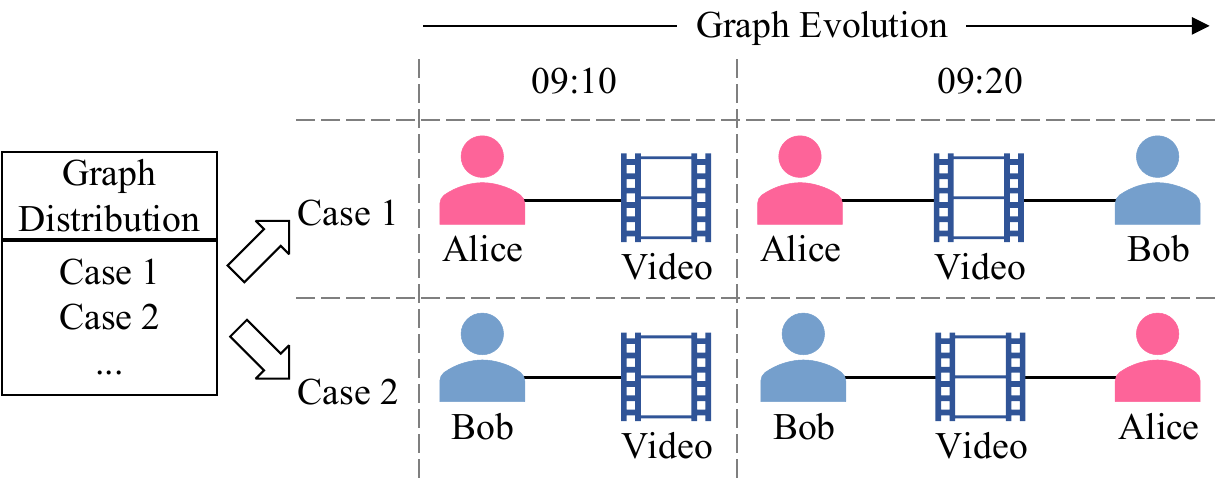} 
\caption{\change{
A toy example of the evolution and the distribution of graphs. In detail, suppose a new video is uploaded at 09:00. Next, different observed graphs, such as Case 1 and Case 2, are sampled from a
graph distribution and then evolve respectively. For instance, in Case 1 (Case 2), Alice (Bob) watches the video at 09:10 and Bob (Alice) watches it at 09:20. If the observed graph is obtained at 09:15 (09:17), the watching behavior from Bob (Alice) is unseen.
}}
\label{fig:example}
\end{figure}

However, the evolution and the distribution of graphs remains unconsidered in existing view generation strategies.
Intuitively, a graph is formed with nodes and edges created in succession, and the observed graph is a snapshot of this procedure.
Thus, the non-connected node pairs are possible to form new edges in future.
\change{Case 1 in Figure~\ref{fig:example} demonstrates this intuition. Furthermore, the evolution of graphs varies (e.g., Case1, Case2, or other possible cases in Figure~\ref{fig:example}) based on the distribution of graphs. Then, exploring the distribution of graphs helps to search unseen but should existing edges, which benefits the variety of generated views and boosts the performance of GCL.}

\begin{figure}[ht]
\centering
\includegraphics[width=0.8\linewidth]{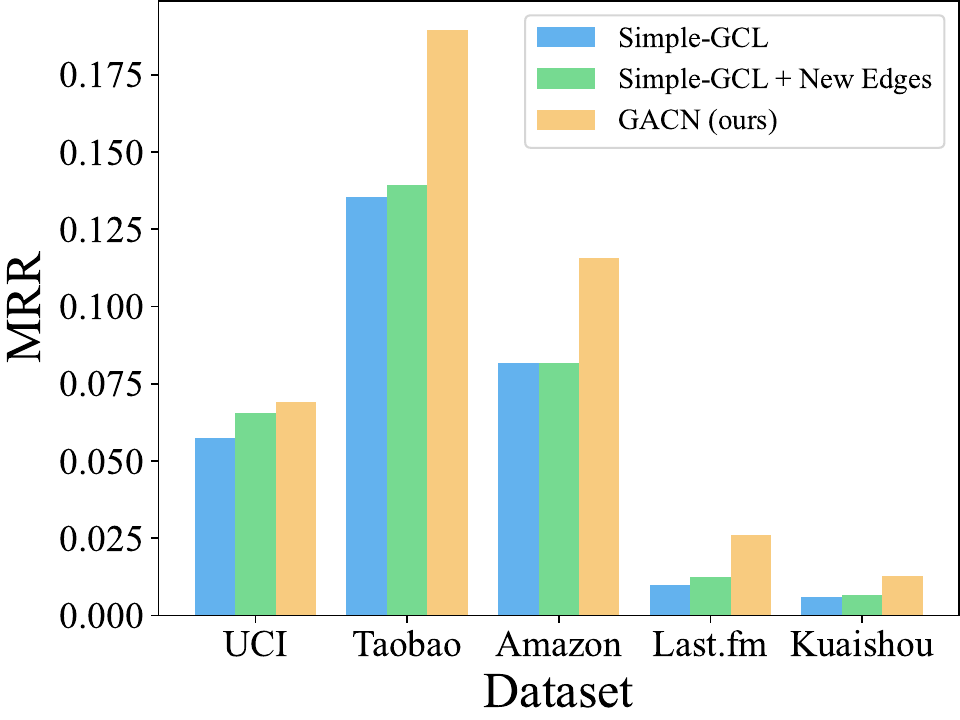} 
\caption{
Empirical experiments show that replacing some existing edges with random new edges in one of the augmented views can improve Simple-GCL on most datasets.
However, it requires a \change{trial-and-error} selection of the new edge rate to get the best performance.
In contrast, \change{our} GACN can automatically learn the graph distribution and precisely add edges for better graph representation learning.}
\label{fig:motivation}
\end{figure}

As illustrated in Figure~\ref{fig:motivation}, we consider replacing some existing edges with new edges randomly in one of the augmented views of a graph contrastive learning method (i.e., Simple-GCL in Sec~\ref{sec:baseline}) and evaluate the link prediction performance.
It is observed that replacing a certain amount of existing edges with new edges can improve Simple-GCL on most datasets, showing the benefit of the supplement of unseen edges.
However, different rates of new edges are required to get the best performance on different datasets due to different graph distributions.



In this work, we argue that the process of data augmentation for GCL should systematically consider graph evolution, and then propose to leverage graph generative adversarial networks (GANs) for graph distribution learning.
Clearly, it is not a trivial task considering the following two major challenges:

\noindent\textbf{Automatically capturing the characteristic of graphs for augmentations.}
On the one hand, compared to image data and text data, graph data are more abstract~\citep{suresh2021adversarial}, making it hard to characterize the distribution of graph.
On the other hand, graph data are discrete (e.g., the value of the adjacent matrix is binary), and sampling-based graph generators are usually hard to be trained end-to-end.
Thus, more explorations are required to design a graph GAN to generate high-quality augmented views.

\noindent\textbf{Jointly training the graph GAN model and the GCL model.}
A simple idea to combine graph GANs with GCL is to train the two models separately.
However, in this way, the connection between the two models is weak and there is no guarantee that the generated views, which deceive the discriminator of the GAN model well, can be well encoded by the GCL model.
Furthermore, training graph GAN and GCL separately needs to maintain two groups of GNN parameters, which is unnecessary.
Thus, it is better to explore a parameter sharing strategy and a jointly learning framework for the sake of effectiveness and efficiency.

To tackle the above challenges, this paper proposes GACN, a graph \textbf{G}enerative \textbf{A}dversarial \textbf{C}ontrastive learning \textbf{N}etwork.
Specifically, GACN develops a graph generative adversarial network with a view generator and a view discriminator to learn generating augmented views through a minimax game. Then, GACN adapts a GNN as the graph encoder and designs two self-supervised learning losses to optimize the parameters. To train GACN, a jointly learning framework is proposed to optimize the view generator, the view discriminator and the graph encoder sequentially and iteratively.

The main contributions of this paper are highlighted as follows:
\begin{itemize}[leftmargin=*]
    \item We explore the benefit of leveraging \change{unseen} edges to boost GCL, and first propose to incorporate graph GANs to learn and generate views for GCL. 
    \item We present GACN, a new graph neural network that develops a view generator and a view discriminator to learn generating views for the graph encoder. All these modules are trained jointly with a novel framework (Section~\ref{sec:methods}).
    \item We conduct comprehensive experiments to evaluate GACN with twelve state-of-the-art baseline methods. The experimental results show that GACN is superior to other methods, and is able to generate views satisfying the famous preferential attachment rule (Section~\ref{sec:exp}).
\end{itemize}


\section{Related Work}
\label{sec:related_work}
In this section, the related work to this study is briefly summarized, including graph contrastive learning and graph generative adversarial network.

\subsection{Graph Contrastive Learning}
Contrastive Learning (CL)~\citep{linsker1988self,van2018representation,tian2020contrastive,1992Self,henaff2020data,hjelm2018learning} is an emerging paradigm to learn quality discriminative representations based on augmented ground-truth samples.
It initially showed the promising capability in the field of computer vision (CV) and natural language processing (NLP) while recently researchers have applied CL to graph domains to fully exploit graph structure information and rich unlabeled data.
The core idea of GCL is to maximize the mutual information between instances (e.g., node, subgraph, and graph) of different views augmented from the original graph.

Similar to the visual domain ~\citep{chen2020simple, tian2020contrastive}, there are various augmentation techniques on attributes or topologies and contrastive pretext tasks on different granularities.
For example, DGI ~\citep{velickovic2019deep} performs the row-wise shuffling on the attribute matrix and conducts node-graph level contrast while MVGRL~\citep{hassani2020contrastive} applies an edge diffusion augmentation to obtain contrasting views.
On top of attribute masking, GraphCL ~\citep{you2020graph} proposes several topology-based augmentations including node dropout, edge dropout and subgraph sampling to incorporate various priors.
Rather than contrasting views at the graph level, GRACE ~\citep{zhu2020deep}, GCA ~\citep{zhu2021graph} and GROC ~\citep{jovanovic2021towards} conduct node-level same-scale contrast, which is the most adopted method to learn node-level representations.
Very recently, JOAO~\citep{you2021graph} adopts a bi-level optimization framework to learn graph data augmentations.

However, many GCL methods require a trial-and-error selection or domain knowledge to augment views, which limits the application of GCL.

\begin{figure*}[ht]
\centering
\includegraphics[width=0.95\linewidth]{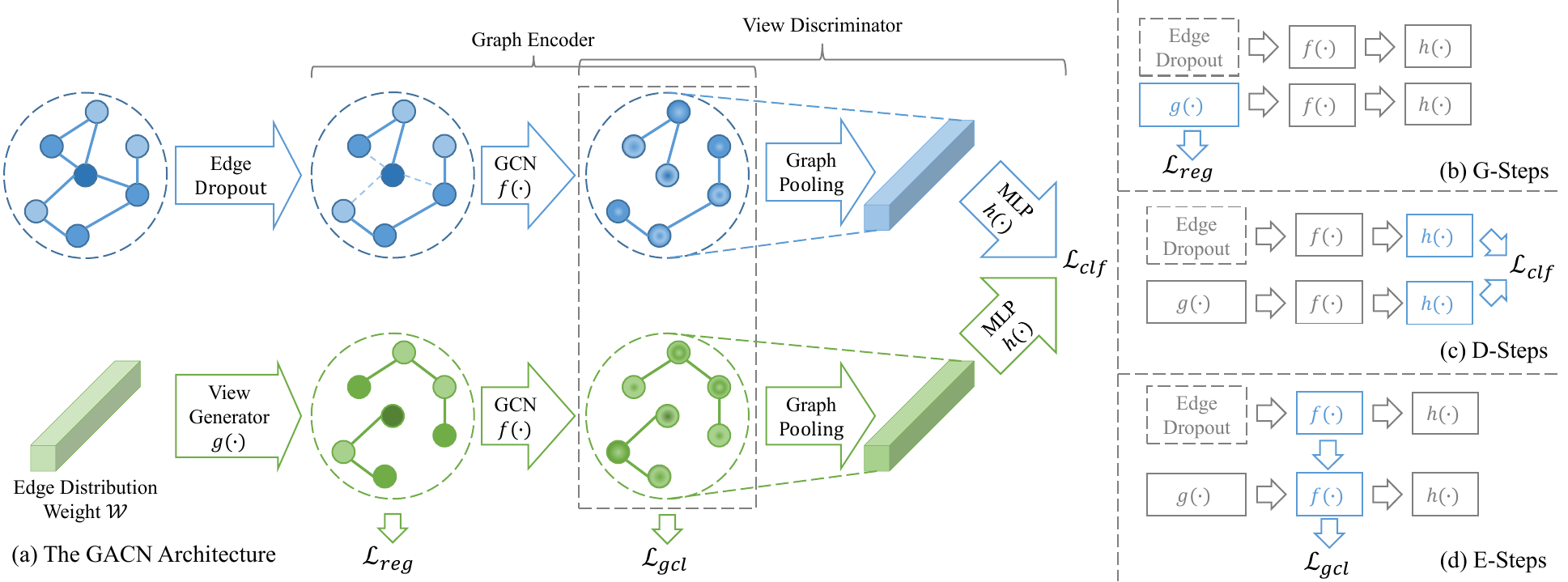} 
\caption{\change{The architecture and the training steps of GACN. There are three modules in GACN, including the view generator, the view discriminator and the graph encoder, which are optimized by the G-Steps, the D-Steps and the E-Steps, respectively.}}
\label{fig:architure}
\end{figure*}

\subsection{Graph Generative Adversarial Network}
By designing a game-theoretical minimax game, Generative Adversarial Networks (GAN)~\citep{goodfellow2014generative} have achieved success in various applications, such as image generation~\citep{denton2015deep}, sequence generation~\citep{yu2017seqgan}, dialogue generation~\citep{li2017adversarial}, information retrieval~\citep{wang2017irgan}, and domain adaption~\citep{zhang2017aspect}. More recently, GANs have been applied on graph-based tasks.

In terms of the graph generation task, ~\citet{liu2019learning} stack multiple GANs to form a hierarchical architecture for preserving topological features of training graphs. 
To preserve the distribution of links with minimal risk of privacy breaches,  ~\citet{tavakoli2017learning} utilize GANs to learn the probability of link formation. 
Aiming at better capturing the essential properties and preserving the patterns of real graphs, \citet{bojchevski2018netgan} introduce NetGAN to learn a distribution of network via the random walks.

Another line of applications is graph embedding.
ANE ~\citep{dai2018adversarial} treats GANs as an regularization term to learn robust representations.
GraphGAN ~\citep{wang2018graphgan} designs a generator to learn node embeddings and a discriminator to predict link probabilities.
NetRA~\citep{yu2018learning} and ProGAN~\citep{gao2019progan} preserve and learn the underlying node similarity in the model of GAN.
Besides, \citet{lei2019gcn} and \citet{yang2019advanced} combine GANs with various encoders
to refine the performance of temporal link prediction.
\citet{sun2019megan} introduce MEGAN for multi-view network embedding, which accounts for the information from individual views and correlations among different views.

Recently, GASSL~\citep{yang2021graph} and AD-GCL~\citep{suresh2021adversarial} incorporate adversarial learning with graph contrastive learning to avoid capturing redundant information by optimizing adversarial graph augmentation strategies.
\change{In computer vision, researchers have tried to make a combination of GAN and CL to boost the performance of GAN~\citep{lee2021infomax} or CL~\citep{pan2021videomoco}.}
However, \change{in graph mining}, adapting GANs to learn graph distribution and generate views for GCL remains unexplored.

\begin{table}[t]
\footnotesize
\centering
\caption{Notations in this paper.}
\begin{tabular}{c|l}
\toprule
Notations & Definitions \\
\midrule
$\mathcal{V}$ & the set of nodes of a graph \\
$\mathcal{E}$ & the set of edges of a graph, $\mathcal{E}\subseteq
\mathcal{V}\times\mathcal{V}$\\
$\mathcal{G}$ & a graph $\mathcal{G}= (\mathcal{V}, \mathcal{E})$ \\
$\mathcal{X}$ & the node attributes that $\mathcal{G}$ may have \\
$F$ & the dimension of node attributes \\
$D$ & the dimension of node representations \\
$A$ & the adjacent matrix of $\mathcal{G}$, $A\in\mathbb{R}^{|\mathcal{V}|\times|\mathcal{V}|}$ \\
$P$ & the approximation of the generated adjacent matrix\\
$d_v$ & the node representation of node $v\in\mathcal{V}$ \\
$\theta_g$ & the parameters of the view generator  \\
$\theta_h$ & the parameters of the view discriminator  \\
$\theta_f$ & the parameters of the graph encoder  \\
\bottomrule
\end{tabular}
\label{tab:notations}
\end{table}

\section{Preliminaries}
\label{sec:preliminaries}
In this section, we introduce some preliminary concepts and notations.
In this work, we denote a graph as $\mathcal{G} = (\mathcal{V}, \mathcal{E})$, where $\mathcal{V}$ is a node set and $\mathcal{E}$ is an edge set.
$\mathcal{G}$ may have node attributes $\{\mathcal{X}_v \in \mathbb{R}^F | v \in \mathcal{V}\}$. The adjacent matrix of $\mathcal{G}$ is denoted as $A\in\mathbb{R}^{|\mathcal{V}|\times|\mathcal{V}|}$, where:
\begin{align}
    A_{i,j} = \left\{
         \begin{array}{lr}
         1, \quad \mbox{if $(v_i, v_j)\in\mathcal{E}$;} \\
         0, \quad \mbox{otherwise.}
         \end{array}
    \right.
\end{align}

\noindent \underline{\textbf{Graph Representation Learning.}}
Given a graph $\mathcal{G} = (\mathcal{V}, \mathcal{E})$,  the aim of graph representation learning is to learn an encoder $f: \mathcal{V}\to\mathbb{R}^D$, where $\{f(v)|v\in\mathcal{V}\}$ can be further used in downstream tasks, such as node classification and link prediction.

\noindent \underline{\textbf{Graph Neural Networks (GNNs).}}
In this work, we focus on using GNNs as the encoder $f$. For a graph $\mathcal{G} = (\mathcal{V}, \mathcal{E})$, each node $v\in\mathcal{V}$ is paired with a representation $d_v$ initialized as $d_v^{(0)} = \mathcal{X}_v$.
The idea is to apply the neighborhood aggregation scheme on $\mathcal{G}$, updating the representation of node by aggregating the representations of neighbor nodes:
\begin{align}
    d_v^{(l)} = AGG(d_v^{(l-1)}, \{d_u^{(l-1)}|u\in \mathcal{N}_v\}),
\end{align}
where $d_v^{(l)}$ denotes the representation of node $v$ in the $l$-th layer, $\mathcal{N}_v$ is the set of neighbors of node $v$, and $AGG$ is the aggregation function. After obtaining $L$ layers presentations, a readout function is adopted to generate the final representation of node $i$:
\begin{align}
    d_v = readout(\{d_v^{(l)}|l=\{0,\cdots,L\}\}).
\end{align}

\noindent \change{\underline{\textbf{Graph Contrastive Learning (GCL).}}
GCL aims to maximize the mutual information between instances (e.g., node, subgraph and graph) of different views augmented from the original graph. Typically, GCL methods adopt graph augmentation strategies to construct positive pairs and negative pairs, and utilize GNNs to encode them into representations. Then, a contrastive loss function is defined to enforce maximizing the consistency between positive pairs compared with negative pairs.}

\change{In this paper, we use LightGCN~\citep{he2020lightgcn} as the GNN encoder and focus on node-level GCL.}
Note that several important notations used in this paper are summarized in Table~\ref{tab:notations}.

\section{Methods}
\label{sec:methods}

In this section, we first present the overview of the proposed GACN model (see Figure~\ref{fig:architure}), and then bring forward the details of its three modules. 
Finally, we propose the optimization framework of GACN.

\subsection{Overview}
As shown in Figure~\ref{fig:architure}a, GACN consists of three modules, namely \textit{View Generator}, \textit{View Discriminator} and \textit{Graph Encoder}.
Specifically, the view generator learns the distributions of edges and generates augmented views by edge sampling.
Then the view discriminator is designed to distinguish views generated by the generator from those generated by predefined augmentation strategies (e.g., edge dropout).
The view generator and the view discriminator are trained in an adversarial style to generate high-quality views.
These views are used to train robust node representations in the graph encoder, which shares the same node representations with the view discriminator.

Note that we do not explicitly encode any graph generative principles into the model design.
However, surprisingly our proposed GACN learns the graph distribution to generate views that follow the well-known preferential attachment rule~\citep{barabasi1999emergence} (see Section~\ref{sec:case}).

\subsection{View Generator}
\label{sec:generator}
Given a graph $\mathcal{G} = (\mathcal{V}, \mathcal{E})$, the view generator is designed to generate a set of augmented views. For a specific view $\mathcal{G}_g$, we assume that each edge $(v_i, v_j)$ in $\mathcal{G}_g$ is associated with a random variable $P_{i,j}\sim\mbox{Bernoulli}(\mathcal{W}_{i,j})$, where $\mathcal{W}\in\mathbb{R}^{|\mathcal{V}|\times|\mathcal{V}|}$ is a learnable matrix, $P$ is a binary matrix with size $|\mathcal{V}|\times|\mathcal{V}|$, $(v_i, v_j)$ is in $\mathcal{G}_g$ if $P_{i,j} = 1$ and is dropped otherwise. To train the view generator in an end-to-end fashion, we relax the discrete $P_{i,j}$ to be a continuous variable in $(0, 1)$ as follows:
\begin{align}
\label{eq:gen}
    P = \sigma(\frac{\mathcal{W} - X_g}{\tau_g}),
\end{align}
where $X_g\in\mathbb{R}^{|\mathcal{V}|\times|\mathcal{V}|}$ is a random matrix with each element sampled from a uniform distribution: $X_{i,j}\sim U(0, 1)$, $\sigma(x) = \frac{1}{1 + e^{-x}}$ is the sigmoid function, and $\tau_g\in(0,1]$ is a hyper-parameter to make $P_{i,j}$ close to $0$ or $1$. Here, $P$ can be treated as an approximation of the generated adjacent matrix.

To constrain the structure of the generated views, we propose two regularization losses, namely the \textbf{Edge Count Loss} and the \textbf{New Edge Loss}, to train the parameters of the generator, i.e., $\Theta_g = \{\mathcal{W}\}$.

\noindent \underline{\textbf{Edge Count Loss.}} This loss is designed to limit the number of edges in $\mathcal{G}_g$. Inspired by the edge-dropout strategy~\citep{zhu2020deep}, we set a ratio $\lambda_g$ and train $\mathcal{W}$ to generate views with $\lambda_g\cdot|\mathcal{E}|$ edges. Formally, the edge count loss is computed as:
\begin{align}
\mathcal{L}_{cnt} = |\lambda_g\cdot|\mathcal{E}| - \sum_{i,j}{P_{i,j}}|.
\end{align}

\noindent \underline{\textbf{New Edge Loss.}} This loss is proposed to avoid generating views that are aggressively different from $\mathcal{G}$. Specifically, we calculate a penalty for each new edge in $\mathcal{G}_g$, i.e., edge $(v_i, v_j)$ with $A_{i,j} = 0$ and $P_{i,j} = 1$. Then, the new edge loss is the sum of all the penalties:
\begin{align}
\mathcal{L}_{new} = \sum_{i,j}(1 - A_{i,j})\cdot P_{i,j}
\end{align}

Then, the regularization loss is the combination of the above two losses:
\begin{align}
\label{eq:reg}
    \mathcal{L}_{reg} = \lambda_{cnt}\cdot\mathcal{L}_{cnt} + \lambda_{new}\cdot\mathcal{L}_{new},
\end{align}
where $\lambda_{cnt}$ and $\lambda_{new}$ are hyper-parameters to balance the influences of $\mathcal{L}_{cnt}$ and $\mathcal{L}_{new}$, respectively.

Besides, for the sake of efficiency, we initialize $\mathcal{W}$ instead of training from scratch. Specifically, we set an initialization rate $\gamma \in [0, 1]$ to constrain the number of new edges at the beginning, i.e., $\gamma\cdot\lambda_{g}\cdot|\mathcal{E}|$ new edges and $(1-\gamma)\cdot\lambda_{g}\cdot|\mathcal{E}|$ existing edges are expected in the generated views. Thus, $\mathcal{W}$ is initialized as follows:
\begin{align}
    \mathcal{W}_{i,j} = \left\{
         \begin{array}{cl}
         \frac{(1-\gamma)\cdot\lambda_{g}\cdot|\mathcal{E}|}{|\mathcal{E}|}&, \quad \mbox{if $(v_i, v_j) \in \mathcal{E}$;} \\
         \frac{\gamma\cdot\lambda_{g}\cdot|\mathcal{E}|}{|\mathcal{C}|}&, \quad \mbox{if $(v_i, v_j) \in \mathcal{C}$;} \\
         0 &, \quad \mbox{otherwise,}
         \end{array}
    \right.
\end{align}
where $\mathcal{C}\subseteq (\mathcal{V}\times\mathcal{V} - \mathcal{E})$ is a candidate set of new edges. Note that we do not consider all the possible new edges as candidates because maintaining a dense $\mathcal{W} \in \mathbb{R}^{|\mathcal{V}|\times|\mathcal{V}|}$ for large graphs is memory-unfriendly. In our implementation, we choose edges related to nodes with top-$2,000$ degrees as the candidate set. Note that we also try top-$5,000$ and top-$10,000$. However, the performance gain of top-$5,000$ or top-$10,000$ over that of top-$2,000$ is almost nothing with a huge increment of training time.

\subsection{View Discriminator}
\label{sec:discriminator}
The view discriminator is a graph-level classifier to  recognize the generated views.
More precisely, the discriminator takes an adjacent matrix as input and judges whether the matrix is a true matrix (i.e., a matrix generated by predefined augmentation strategies) or a false matrix (i.e., a matrix generated by the view generator).
Formally, given a graph $\mathcal{G} = (\mathcal{V}, \mathcal{E})$, we denote the set of views generated by predefined augmentation strategies (i.e., edge dropout in this work) as $\vec{\mathbf{\mathcal{G}}}_p$, and the set of views generated by the view generator as $\vec{\mathbf{\mathcal{G}}}_g$. Thus, for each $G \in \vec{\mathbf{\mathcal{G}}}_p\cup\vec{\mathbf{\mathcal{G}}}_g$, a GNN encoder $f$ is used to encode the representations of each node:
\begin{align}
\label{eq:enc}
    \{d_v|v\in\mathcal{V}\} = f(G).
\end{align}

Then, we calculate the graph representation by concatenating the mean pooling and the maximum pooling of the node representations:

\begin{align}
\label{eq:pool}
    d_G = (\frac{1}{|\mathcal{V}|}\sum_{v\in\mathcal{V}}{d_v}) \oplus \mbox{MaxPool}(\{d_v|v\in\mathcal{V}\}),
\end{align}
where $\oplus$ is the concatenate operation. With the graph representation, we compute the probability of $G$ using an $L_h$-layer Multilayer Perceptron (MLP) $h$:
\begin{align}
\label{eq:mlp}
    p_G = h(d_G),
\end{align}
where $\Theta_h = \{(W_i, b_i)|i = 1,2,\ldots,L_h\}$ is the set of parameters in $h$.

To train the discriminator, we label the views in $\vec{\mathbf{\mathcal{G}}}_p$ with $1$ and the views in $\vec{\mathbf{\mathcal{G}}}_g$ with $0$. Suppose that the label of $G$ is $y_G$. Then, the classification loss is defined as follows:
\begin{align}
\label{eq:clf}
    \mathcal{L}_{clf}=-y_G\cdot\log(p_G)-(1-y_G)\cdot\log(1-p_G).
\end{align}

\subsection{Graph Encoder}
\label{sec:encoder}
The graph encoder is designed to learn the node representations, i.e., the set of parameters of the encoder is $\Theta_f = \{d_v^{(0)}|v\in\mathcal{V}\}$, and is trained by two self-supervised losses, including the \textbf{Graph Contrastive Loss} and the pairwise \textbf{Bayesian Personalized Ranking (BPR) Loss}.

\noindent \underline{\textbf{Graph Contrastive Loss.}}
This loss is proposed to learn robust node representations through maximizing the agreement between different views of the same node compared to that of other nodes.
Specifically, we generate two views $\mathcal{G}_p$ and $\mathcal{G}_g$ using the predefined augmentation strategies and the view generator, respectively. Encoding $\mathcal{G}_p$ and $\mathcal{G}_g$, we have two set of node representations:
\begin{align}
    \{d_v^p|v\in\mathcal{V}\} = f(\mathcal{G}_p),
    \notag \\
    \{d_v^g|v\in\mathcal{V}\} = f(\mathcal{G}_g).
\end{align}
Then the graph contrastive loss is defined as :
\begin{align}
\label{eq:gcl}
    \mathcal{L}_{gcl} = -\sum_{v\in\mathcal{V}}\log{\frac{\exp(\frac{{d_v^p}^\top d_v^g}{\tau_f})}{\sum_{u\in\mathcal{V}}{\exp(\frac{{d_u^p}^\top d_v^g}{\tau_f})}}},
\end{align}
where $\tau_f$ is the temperature hyper-parameter in softmax.

\noindent \underline{\textbf{Bayesian Personalized Ranking Loss.}}
This loss is introduced to learn representations that are suitable for downstream tasks, especially for link prediction, and the intuition is to maximize the similarity of connected nodes, while minimize the similarity of disconnected nodes.
Formally, the bpr loss is defined as:
\begin{align}
\label{eq:bpr}
    \mathcal{L}_{bpr} = -\frac{1}{|\mathcal{O}|}\sum_{(i,j,k)\in\mathcal{O}}{\log\sigma(d_i^\top d_j - d_i^\top d_k)},
\end{align}
where $\mathcal{O} = \{(i,j,k)|A_{i,j} = 1, A_{i,k} = 0\}$ is the training data.

Then, the self-supervised loss is the combination of the above two losses:
\begin{align}
\label{eq:ssl}
    \mathcal{L}_{ssl} = \lambda_{gcl}\cdot\mathcal{L}_{gcl} + \lambda_{bpr}\cdot\mathcal{L}_{bpr},
\end{align}
where $\lambda_{gcl}$ and $\lambda_{bpr}$ are hyper-parameters to balance the influences of $\mathcal{L}_{gcl}$ and $\mathcal{L}_{bpr}$, respectively.

\begin{algorithm}[t]
\footnotesize
\caption{GACN Framework.}
\label{alg:training_framework}
\LinesNumbered 
\KwIn{
graph $\mathcal{G} = (\mathcal{V}, \mathcal{E})$, dimension of embedding $s$, hyper-parameters $\tau_g, \tau_f, \lambda_{g}, \lambda_{cnt}, \lambda_{new}, \lambda_{gcl}, \lambda_{bpr}$
}
\KwOut{node representations $\{d_v|v\in\mathcal{V}\}$}
Initialize $\Theta_g, \Theta_h, \Theta_f$\;
\While{GACN not converge}{
    \For{G-Steps}{
        Sample $X_g$ and calculate $P$ according to Eq.~(\ref{eq:gen}) \;
        Compute $p_{\mathcal{G}_g}$ using $P$ according to Eq.~(\ref{eq:enc}), (\ref{eq:pool}) and (\ref{eq:mlp})\;
        Update $\Theta_g$ according to Eq.~(\ref{eq:reg}) and (\ref{eq:clf_g})\;
    }
    \For{D-Steps}{
        Generate and label $\vec{\mathbf{\mathcal{G}}}_p, \vec{\mathbf{\mathcal{G}}}_g$ \;
        Encode $G \in \vec{\mathbf{\mathcal{G}}}_p\cup\vec{\mathbf{\mathcal{G}}}_g$ using Eq.~(\ref{eq:enc}), (\ref{eq:pool}) and (\ref{eq:mlp})\;
        Update $\Theta_h$ according to Eq.~(\ref{eq:clf})\;
    }
    \For{E-Steps}{
        Generate $\mathcal{G}_p, \mathcal{G}_g$ for Eq.~(\ref{eq:gcl}) and sample $\mathcal{O}$ for Eq.~(\ref{eq:bpr}) \;
        Update $\Theta_f$ according to Eq.~(\ref{eq:ssl})\;
    }
}
$\{d_v|v\in\mathcal{V}\} = f(\mathcal{G})$ \;
\Return{$\{d_v|v\in\mathcal{V}\}$}.
\end{algorithm}

\subsection{Model Optimization}
\label{sec:optimization}
In this subsection, we present the parameter optimization procedure of GACN. As shown in Algorithm~\ref{alg:training_framework}, the view generator, the view discriminator and the graph encoder are optimized sequentially and iteratively.

\underline{\textbf{G-Steps (Lines 3--7)}} (see Figure~\ref{fig:architure}b) optimize the parameters of the view generator. Specifically, in each iteration, an augmented view $\mathcal{G}_g$ is generated and then the regularization loss is computed. In consideration of generating high-quality views, an adversarial classification loss is incorporated to cheat the view discriminator by labeling $\mathcal{G}_g$ with $1$. According to Eq.~(\ref{eq:clf}), we have:
\begin{align}
\label{eq:clf_g}
    \mathcal{L}'_{clf}=-\log(p_{\mathcal{G}_g}).
\end{align}

\underline{\textbf{D-Steps (Lines 8--12)}} (see Figure~\ref{fig:architure}c) optimize the parameters of the view discriminator by generating $\vec{\mathbf{\mathcal{G}}}_p, \vec{\mathbf{\mathcal{G}}}_g$ and training the discriminator to classify them.

\underline{\textbf{E-Steps (Lines 13--17)}} (see Figure~\ref{fig:architure}d) first prepare the training data for the self-supervised losses and then update the parameters of the graph encoder.

Note that all the parameters are optimized using the back propagation algorithm. After converging, we obtain the learned node representations $\{d_v|v\in\mathcal{V}\}$ by encoding graph $\mathcal{G}$ (Lines 18--19).

\underline{\textbf{Time Complexity Analysis.}}
For the view generator, the time complexity to generate a single view and calculate the regularization loss is $O(|\mathcal{V}|^2)$.
For the view discriminator, the time complexity to encode a graph using LightGCN~\citep{he2020lightgcn} is $O(|\mathcal{V}|^2\cdot N_lD)$, where $N_l$ is the number of GCN layers. The time complexity to pool the graph and calculate $\mathcal{L}_{clf}$ is $O(|\mathcal{V}|\cdot D)$.
For the graph encoder, the time complexity to encode the two augmented views is $O(|\mathcal{V}|^2\cdot N_lD)$ and the time complexity to compute $\mathcal{L}_{ssl}$ is $O(|\mathcal{V}|^2\cdot D) + |\mathcal{O}|\cdot D)$.
Thus, the overall time complexity of Algorithm~\ref{alg:training_framework} is
$O(N_{iter}[N_G + N_D\cdot(|\vec{\mathbf{\mathcal{G}}}_p| + |\vec{\mathbf{\mathcal{G}}}_g|) + N_E]\cdot|\mathcal{V}|^2\cdot N_lD)$
, where $N_{iter}$ is the number of iteration round, $N_G$, $N_D$ and $N_E$ are the number of G-Steps, D-Steps and E-Steps, respectively.
\section{Experiments}
\label{sec:exp}
In this section, we conduct extensive experiments and answer the following research questions:
\begin{itemize}[leftmargin=*]
    \item \textbf{RQ1}: How does GACN perform w.r.t. node classification task?
    \item \textbf{RQ2}: How does GACN perform w.r.t. link prediction task?
    \item \textbf{RQ3}: What are the benefits of the proposed modules of GACN?
    \item \textbf{RQ4}: Can the generator of GACN generate high-quality graphs for contrastive learning?
    \item \textbf{RQ5}: How do different settings influence the effectiveness of GACN?
\end{itemize}

\begin{table}
\footnotesize
\centering
\caption{The statistics of datasets.}
\begin{tabular}{cccc}
\toprule
Datasets & $|\mathcal{V}|$ & $|\mathcal{E}|$ & Task \\
\midrule
Cora & 2,708 & 5,429 & Node Classification\\
Citeseer & 3,312 & 4,714 & Node Classification\\
\midrule
UCI & 1,677 & 56,617 & Link Prediction \\
Taobao & 12,611 & 20,890 & Link Prediction \\
Amazon & 10,099 & 148,659 & Link Prediction \\
Last.fm & 127,786 & 720,537 & Link Prediction \\
Kuaishou & 138,812 & 1,779,639 & Link Prediction \\
\bottomrule
\end{tabular}
\label{tab:datasets}
\end{table}

\subsection{Experimental Settings}
\subsubsection{Datasets}
We evaluate the performance of GACN on seven real-world datasets, including two datasets for node classification namely Cora and Citeseer, and five datasets for link prediction namely UCI, Taobao, Amazon, Lastfm and Kuaishou. We summarize the statistics of all the datasets in Table \ref{tab:datasets}. The detailed information of these datasets is listed as follows.

\noindent \textbf{Datasets for Node Classification.}
\begin{itemize}[leftmargin=*]
    \item \textbf{Cora}~\citep{mccallum2000automating} consists of 2,708 scientific publications classified into one of seven classes. The citation network consists of 5,429 edges. Each publication in the dataset is described by a 0/1-valued word vector indicating the absence/presence of the corresponding word from the dictionary. The dictionary consists of 1,433 unique words.
    \item \textbf{Citeseer}~\citep{giles1998citeseer} is similar to Cora. It  consists of 3,312 nodes and 4,714 edges. The nodes are classified into one of six classes and the dimension of node feature is 3,703.
\end{itemize}

\begin{table}
\footnotesize
\centering
\caption{The experimental results of node classification. The best results are illustrated in bold and the number underlined is the runner-up.}
\begin{tabular}{c|ccc|ccc}
\toprule
Dataset & \multicolumn{3}{c|}{Cora} & \multicolumn{3}{c}{Citeseer} \\
Metric & P & R & F1 & P & R & F1 \\
\midrule
DeepWalk & 0.7753 & 0.7012 & 0.7292 & 0.5579 & 0.4962 & 0.4869 \\
LINE & 0.7873 & 0.6970 & 0.7281 & 0.5992 & 0.4437 & 0.4413 \\
node2vec & 0.7744 & 0.7352 & 0.7516 & 0.4717 & 0.4581 & 0.4552 \\
LightGCN& 0.7615 & 0.7342 & 0.7453 & 0.4434 & 0.4513 & 0.4361 \\
\midrule
Simple-GCL & 0.8491 & 0.8154 & 0.8287 & 0.7019 & 0.6930 & 0.6946 \\
DGI &  0.8320 & 0.8129 & 0.8212 & 0.6427 & 0.6357 & 0.6278 \\
GraphCL & 0.7993 & 0.7500 & 0.7689 & 0.6547 & 0.6156 & 0.6112 \\
GRACE & 0.8546 & \underline{0.8377} & 0.8445 & 0.7232 & 0.6963 & 0.6948 \\
SGL & 0.8029 & 0.7769 & 0.7887 & 0.7177 & 0.7045 & 0.7063 \\
\midrule
GraphGAN & 0.4195 & 0.2177 & 0.1745 & 0.3148 & 0.2994 & 0.2696 \\
AD-GCL & 0.4176 & 0.3672 & 0.3800 & 0.2842 & 0.2809 & 0.2763 \\
GraphMAE & \underline{0.8667} & 0.8287 & \underline{0.8447} & \underline{0.7278} & \underline{0.7048} & \underline{0.7064} \\
\midrule
GACN & \textbf{0.8705} & \textbf{0.8545} & \textbf{0.8614} & \textbf{0.7311} & \textbf{0.7187} & \textbf{0.7212} \\
\bottomrule
\end{tabular}
\label{tab:node_classification}
\end{table}

\noindent \textbf{Datasets for Link Prediction.}
\begin{itemize}[leftmargin=*]
    \item \textbf{UCI}~\citep{konect:2016:opsahl-ucsocial} contains the message communications between the students of the University of California, Irvine in an online community.
    \item \textbf{Taobao}~\citep{zhu2018learning} is offered by Alibaba with user behaviors collected from Taobao\footnote{https://www.taobao.com/}. There are 1,000 users with all the corresponding interactive items in this dataset.
    \item \textbf{Amazon}~\citep{he2016ups} is a network that includes product metadata and links between products. We use the data provided by~\citep{cen2019representation} which contains the product metadata of \textit{Electronic} category.
    \item \textbf{Last.fm}\footnote{https://www.last.fm/} contains $<$user, artist, song$>$ tuples collected from Last.fm API, which represents the whole listening habits of nearly 1,000 users. We use the $<$user, artist$>$ pairs to construct a network.
    \item \textbf{Kuaishou}\footnote{https://www.kuaishou.com/} is collected from the Kuaishou online video-watching platform. This dataset includes the interactions of 6,840 users and 131,972 videos.
\end{itemize}

\subsubsection{Baseline Methods}
\label{sec:baseline}
To demonstrate the effectiveness and efficiency of GACN, we choose twelve state-of-the-art baseline methods, categorized into three groups.
Graph representation learning models include DeepWalk~\citep{perozzi2014deepwalk}, LINE~\citep{tang2015line}, node2vec~\citep{node2vec-kdd2016} and LightGCN~\citep{he2020lightgcn}. 
Graph contrastive learning models include DGI~\citep{velickovic2019deep}, GraphCL~\citep{you2020graph}, GRACE~\citep{zhu2020deep} and SGL~\citep{wu2021self}.
Graph generative and adversarial learning models include GraphGAN~\citep{wang2018graphgan} , AD-GCL~\citep{suresh2021adversarial} and GraphMAE~\citep{10.1145/3534678.3539321}.
The details of the baseline methods are listed as follows.

\begin{table*}
\footnotesize
\centering
\caption{The experimental results of link prediction. The best results are illustrated in bold and the number underlined is the runner-up.}
\begin{tabular}{c|cc|cc|cc|cc|cc}
\toprule
Dataset & \multicolumn{2}{c|}{UCI} & \multicolumn{2}{c|}{Taobao} & \multicolumn{2}{c|}{Amazon} & \multicolumn{2}{c|}{Last.fm} & \multicolumn{2}{c}{Kuaishou}\\
Metric  & H@50 & MRR & H@50 & MRR & H@50 & MRR & H@50 & MRR & H@50 & MRR \\
\midrule
DeepWalk & \underline{0.2550} & 0.0474 & 0.3522 & \underline{0.1764} & \underline{0.4496} & 0.0744 & 0.1344 & 0.0180 & 0.0486 & 0.0055 \\
LINE & 0.1086 & 0.0249 & 0.3117 & 0.1758 & 0.3327 & 0.0661 & 0.0857 & 0.0112 & 0.0223 & 0.0031 \\
node2vec & 0.1808 & 0.0312 & 0.3533 & 0.1754 & 0.3026 & 0.0619 & 0.1184 & 0.0181 & 0.0623 & 0.0073 \\
LightGCN & 0.1093 & 0.0256 & 0.3433 & 0.1692 & 0.4359 & 0.0818 & 0.1368 & \underline{0.0203} & \underline{0.0745} & \underline{0.0093} \\
\midrule
Simple-GCL & 0.2549 & \underline{0.0574} & 0.3330 & 0.1355 & 0.4079 & 0.0818 & 0.0581 & 0.0100 & 0.0511 & 0.0060 \\
DGI & 0.1972 & 0.0310 & 0.1657 & 0.0388 & 0.1463 & 0.0258 & 0.0972 & 0.0151 & 0.0485 & 0.0060 \\
GraphCL & 0.1669 & 0.0291 & 0.1659 & 0.0348 & 0.1692 & 0.0334 & 0.1012 & 0.0145 & 0.0468 & 0.0061 \\
GRACE & 0.1915 & 0.0270 & 0.2006 & 0.1056 & 0.3127 & 0.0553 & \underline{0.1385} & 0.0198 & 0.0439 & 0.0055 \\
SGL & 0.2545 & \underline{0.0574} & \underline{0.3654} & 0.1741 & 0.4014 & 0.0811 & 0.0981 & 0.0150 & 0.0702 & 0.0086 \\
\midrule
GraphGAN & 0.2543 & 0.0374 & 0.3538 & 0.1390 & 0.4380 & \underline{0.0882} & 0.0781 & 0.0115 & 0.0544 & 0.0067 \\
AD-GCL & 0.1819 & 0.0323 & 0.1008 & 0.0214 & 0.0843 & 0.0118 & 0.0822 & 0.0111 & 0.0184 & 0.0024 \\
GraphMAE & 0.0170 & 0.0052 & 0.1366 & 0.0441 & 0.2660 & 0.0348 & 0.0307 & 0.0043 & 0.0206 & 0.0031 \\
\midrule
GACN & \textbf{0.2836} & \textbf{0.0692} & \textbf{0.3794} & \textbf{0.1895} & \textbf{0.5593} & \textbf{0.1158} & \textbf{0.1568} & \textbf{0.0263} & \textbf{0.1067} & \textbf{0.0132} \\
\bottomrule
\end{tabular}
\label{tab:link_prediction}
\end{table*}

\noindent \textbf{Graph Representation Learning Models.}
\begin{itemize}[leftmargin=*]
    \item \textbf{DeepWalk}~\citep{perozzi2014deepwalk} is an embedding method for static homogeneous networks. It exploits the random walk strategy and the skip-gram model to learn node vector representations.
    \item \textbf{LINE}~\citep{tang2015line} learns node representations by modelling the first- and second-order proximity between node pairs.
    \item \textbf{node2vec}~\citep{node2vec-kdd2016} adds two parameters to control the random walk process based on DeepWalk.
    \item \textbf{LightGCN}~\citep{he2020lightgcn} is a light-weight graph convolution network, which is easy to train and has good generalization ability.
\end{itemize}

\noindent \textbf{Graph Contrastive Learning Models.}
\begin{itemize}[leftmargin=*]
    \item \textbf{Simple-GCL} is \change{a variant of SGL-ED~\citep{wu2021self}} implemented by us and is trained using the InfoNCE~\cite{oord2018representation} loss upon two views generated via edge-dropping.
    \item \textbf{DGI}~\citep{velickovic2019deep} relies on maximizing mutual information between patch representations and corresponding high-level summaries of graphs to learn node representations in an unsupervised manner.
    \item \textbf{GraphCL}~\citep{you2020graph} is a graph contrastive learning framework for learning unsupervised representations of graph data, which designs four types of graph augmentations to incorporate various priors.
    \item \textbf{GRACE}~\citep{zhu2020deep} generates two graph views by corruption and learn node representations by maximizing the agreement of node representations in these two views.
    \item \textbf{SGL}~\citep{wu2021self} explores self-supervised learning on graph structure and accordingly devises three unified augmentation operators including node dropout, edge dropout and random walk.
\end{itemize}

\noindent \textbf{Graph Generative and Adversarial Learning Models.}
\begin{itemize}[leftmargin=*]
    \item \textbf{GraphGAN}~\citep{wang2018graphgan} is a graph representation learning framework unifying the generative model and the discriminative model, in which these two models play a game-theoretical minimax game.
    \item \textbf{AD-GCL}~\citep{suresh2021adversarial} proposes a principle to avoid capturing redundant information during the training by optimizing adversarial graph augmentation strategies used in GCL.
    \item \textbf{GraphMAE}~\citep{10.1145/3534678.3539321} explores generative self-supervised learning in graphs and designs a state-of-the-art graph autoencoder using the masked feature reconstruction strategy with a scaled cosine error as the reconstruction criterion. 
\end{itemize}
Note that we focus on node-level tasks in this paper, and methods designed for graph-level tasks such as GCA~\citep{zhu2021graph}, JOAO~\citep{you2021graph}, MVGRL~\citep{hassani2020contrastive} and GASSL~\citep{yang2021graph} are not chosen as baselines. 

\subsubsection{Parameter Settings}
We implement GACN with Pytorch and the model is optimized using the Adam optimizer with learning rate $0.001$ during the training phase.
By default, $\tau_g$ is set to $0.0001$, $\tau_f$ is set to $0.5$, $\lambda_{g}$ is set to $0.5$, $\lambda_{cnt}$ is set to $1$, $\lambda_{new}$ is set to $0.5$, $\gamma$ is set to $0.75$.
For Cora, Citeseer and UCI, $\lambda_{gcl}$ is set to $1$ and $\lambda_{bpr}$ is set to $0.0001$, while for the other datasets, $\lambda_{gcl}$ is set to $0.0001$ and $\lambda_{bpr}$ is set to $1$. 
For all the baseline methods, we tune the parameters according to the validation set and report the best results.
The dimension of embedding $s$ of all the model is set to $128$, and all the experiments are conducted on a single GTX 1080Ti GPU.

\subsubsection{Metrics}
For the node classification task, we choose three widely-used metrics, namely \textit{P(recision)}, \textit{R(ecall)} and \textit{F1}.
For the link prediction task, we adopt two ranking metrics, include \textit{MRR} and \textit{H(it rate)@k}. In this paper, we report \textit{H@50} and similar results are observed when $k=20$ and $k=100$.

\subsection{Node Classification (RQ1)}

We evaluate the performance of GACN w.r.t. the node classification task on the Cora and the Citeseer datasets.
Following the experimental setting as \citet{hassani2020contrastive} and \citet{velickovic2019deep}, we first run different methods without supervision to generate all the nodes' embeddings.
Then we train a linear classifier and report the mean accuracy on the test nodes through 10 random initialization.
As shown in Table~\ref{tab:node_classification}, GACN achieves the best results compared to the SOTA baseline methods in all benchmarks.
Notably, GACN outperforms existing GCL methods by a large margin on two node classification datasets. Notice that AD-GCL is designed for graph-level tasks and has poor performances on node-level tasks.

\subsection{Link Prediction (RQ2)}
In the subsection, we conduct the link prediction on the UCI, Taobao, Amazon, Last.fm and the Kuaishou datasets.
Table~\ref{tab:link_prediction} reports the experimental results and it is observed that:
1) GACN consistently performs the best on all datasets compared to other methods. We attribute these results to the fact that GACN is able to explore unseen edges and generate high-quality views for graph contrastive learning.
2) Although SGL leverages the BPR loss and the contrastive loss for self-supervise learning, GACN still outperforms SGL, showing the effectiveness of incorporating GCL with graph GANs.


\begin{table*}
\footnotesize
\centering
\caption{\change{Ablation Study. The best results are illustrated in bold and the number underlined is the runner-up.}}
\begin{tabular}{c|ccc|ccc|cc|cc|cc|cc|cc}
\toprule
Dataset & \multicolumn{3}{c|}{Cora} & \multicolumn{3}{c|}{Citeseer} & \multicolumn{2}{c|}{UCI} & \multicolumn{2}{c|}{Taobao} & \multicolumn{2}{c|}{Amazon} & \multicolumn{2}{c|}{Last.fm} & \multicolumn{2}{c}{Kuaishou}\\
Metric & P & R & F1 & P & R & F1 & H@50 & MRR & H@50 & MRR & H@50 & MRR & H@50 & MRR & H@50 & MRR \\
\midrule
\textit{w/o} REG & 0.8650 & 0.8385 & 0.8498 & 0.7250 & 0.7136 & 0.7155 & 0.0685 & 0.0127 & 0.3159 & 0.1308 & 0.4850 & 0.0997 & 0.1486 & 0.0230 & 0.0840 & 0.0103  \\
\textit{w/o} GAN & \underline{0.8670} & \underline{0.8497} & \underline{0.8571} & 0.7286 & 0.7170 & 0.7191 & 0.2389 & 0.0627 & 0.3654 & 0.1741 & 0.4025 & 0.0818 & 0.0982 & 0.0150 & 0.0693 & 0.0086 \\
\textit{w/o} SSL & 0.8608 & 0.8368 & 0.8470 & 0.7265 & 0.7151 & 0.7168 & 0.0871 & 0.0160 & 0.1743 & 0.0553 & 0.0380 & 0.0074 &  0.0023 & 0.0004 & 0.0007 & 0.0001 \\
\textit{w/o} GCL & 0.7226 & 0.3442 & 0.3565 & 0.7305 & 0.7175 & 0.7194 & 0.0817 & 0.0171 & \underline{0.3775} & \underline{0.1850} & \underline{0.5064} & \underline{0.1036} & \underline{0.1515} & \underline{0.0247} & \underline{0.1001} & \underline{0.0122}  \\
\textit{w/o} BPR & 0.8632 & 0.8392 & 0.8494 & \underline{0.7307} & \underline{0.7181} & \underline{0.7207} & \underline{0.2727} & \underline{0.0672} & 0.2326 & 0.1270 & 0.3587 & 0.0718 & 0.0497 & 0.0081 & 0.0554 & 0.0066 \\
\midrule
GACN & \textbf{0.8705} & \textbf{0.8545} & \textbf{0.8614} & \textbf{0.7311} & \textbf{0.7187} & \textbf{0.7212} & \textbf{0.2836} & \textbf{0.0692} & \textbf{0.3794} & \textbf{0.1895} & \textbf{0.5593} & \textbf{0.1158} & \textbf{0.1568} & \textbf{0.0263} & \textbf{0.1067} & \textbf{0.0132} \\
\bottomrule
\end{tabular}
\label{tab:ablation}
\end{table*}

\begin{figure}[ht]
  \centering
  \subfigure[$\lambda_{cnt} = 0, \lambda_{new} = 0$]{
    \label{fig:new_0_count_0}
    \begin{minipage}[]{0.475\linewidth}
      \includegraphics[width=1.0\linewidth]{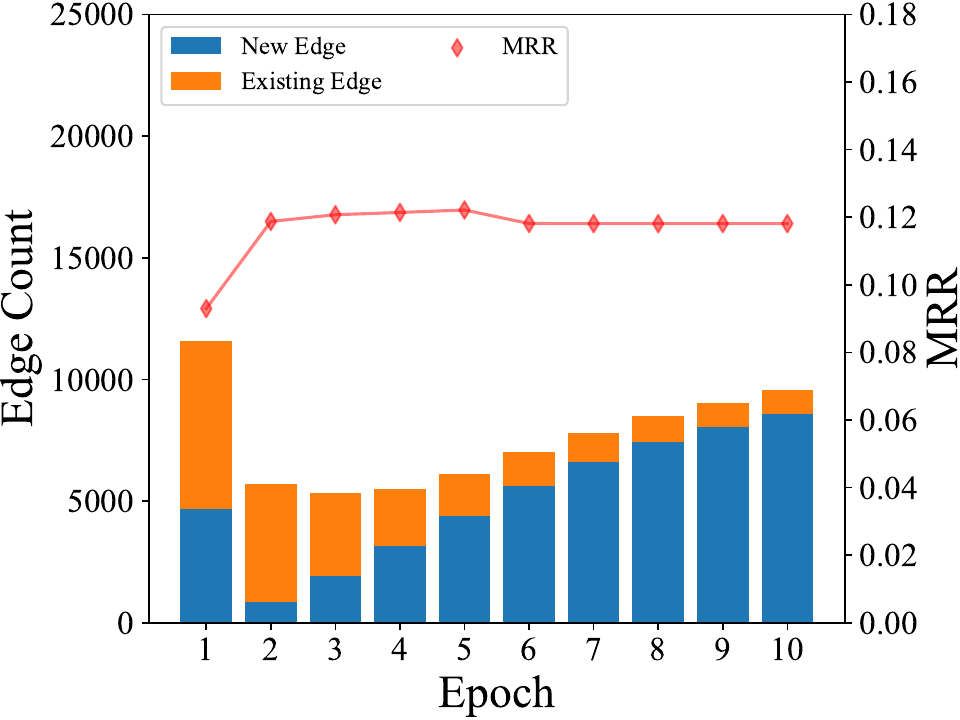}
    \end{minipage}
  }
  \subfigure[$\lambda_{cnt} = 1, \lambda_{new} = 0$]{
    \label{fig:new_0_count_1}
    \begin{minipage}[]{0.475\linewidth}
      \includegraphics[width=1.0\linewidth]{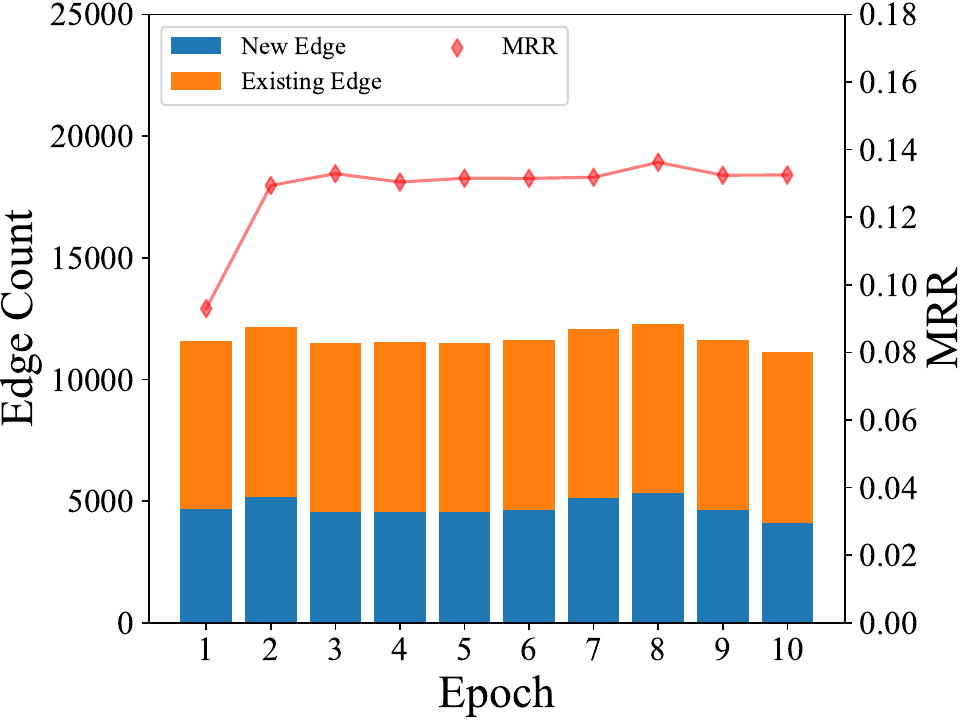}
    \end{minipage}
  }
  \subfigure[$\lambda_{cnt} = 0, \lambda_{new} = 0.5$]{
    \label{fig:new_0_5_count_0}
    \begin{minipage}[]{0.475\linewidth}
      \includegraphics[width=1.0\linewidth]{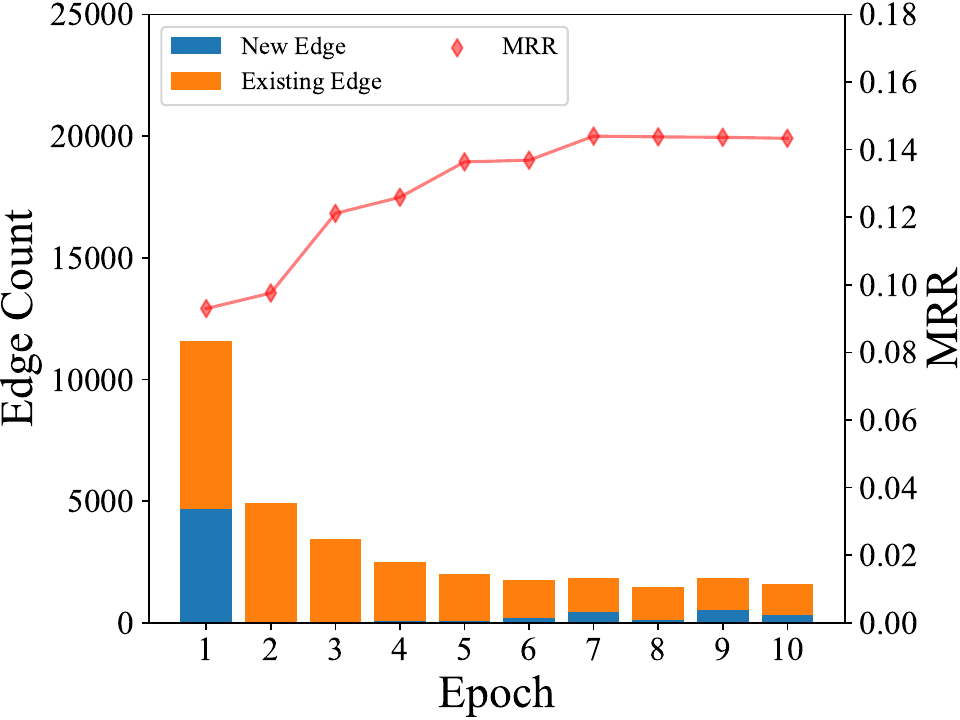}
    \end{minipage}
  }
  \subfigure[$\lambda_{cnt} = 1, \lambda_{new} = 0.5$]{
    \label{fig:new_0_5_count_1}
    \begin{minipage}[]{0.475\linewidth}
      \includegraphics[width=1.0\linewidth]{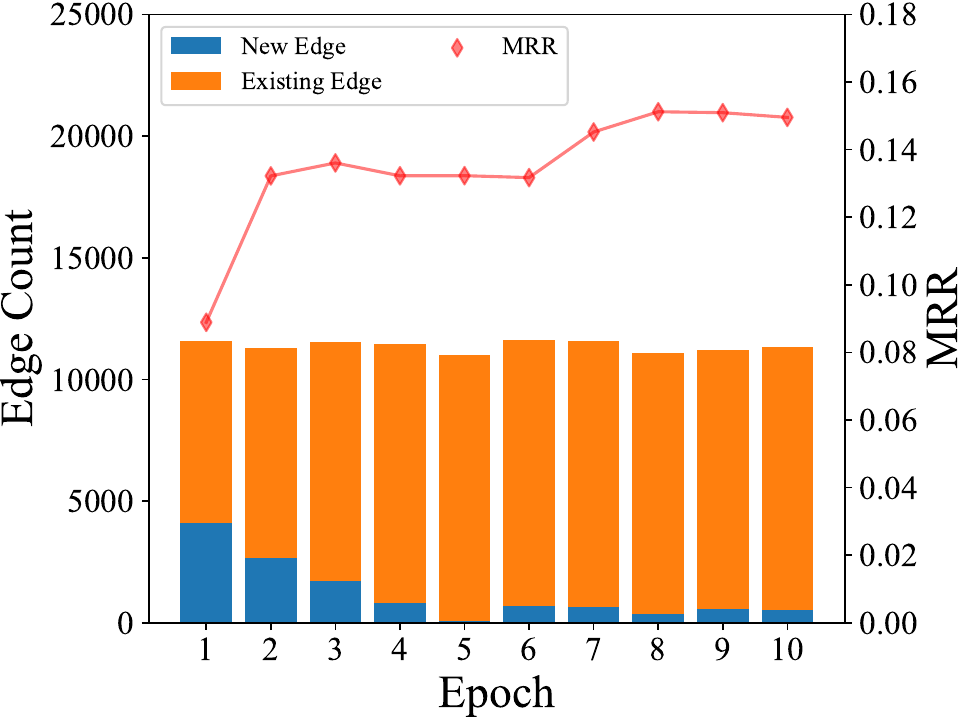}
    \end{minipage}
  }
  \caption{\change{The impact of $\lambda_{cnt}$ and $\lambda_{new}$.}}
  \label{fig:new_count}
\end{figure}

\subsection{Ablation Study (RQ3)}
In this subsection, we evaluate the effectiveness of different components of GACN, with five variants as follows:
\begin{itemize}[leftmargin=*]
    \item \textbf{\textit{w/o} REG}: The view generator is trained without the regularization loss $\mathcal{L}_{reg}$.
    \item \textbf{\textit{w/o} GAN}: The contrastive loss $\mathcal{L}_{gcl}$ is optimized using views generated by predefined augmentation strategies only.
    \item \textbf{\textit{w/o} SSL}: The self-supervised learning losses are ignored and the model is optimized using the G-Steps and D-Steps only.
    \item \textbf{\textit{w/o} GCL}: The graph contrastive loss $\lambda_{gcl}$ is ignored during E-Steps.
    \item \textbf{\textit{w/o} BPR}: The bayesian personalized ranking loss $\lambda_{bpr}$ is ignored during E-Steps.
\end{itemize}

\change{Table~\ref{tab:ablation}} shows the experimental results.
We can find that:
1) The regularization loss plays as an assistant role in performance, which indicates that $\mathcal{L}_{reg}$ helps to generate rational views for contrastive learning.
2) The graph GAN is important in GACN, showing the benefits of utilizing GAN to generate views and the joint learning framework.
3) The self-supervised learning losses are essential to GACN, because the GAN in GACN is based on graph-level classification and does not focus on learning node representations.


\begin{figure}[t]
  \centering
  \subfigure[Random]{
    \label{fig:degree_random}
    \begin{minipage}[]{0.475\linewidth}
      \includegraphics[width=1.0\linewidth]{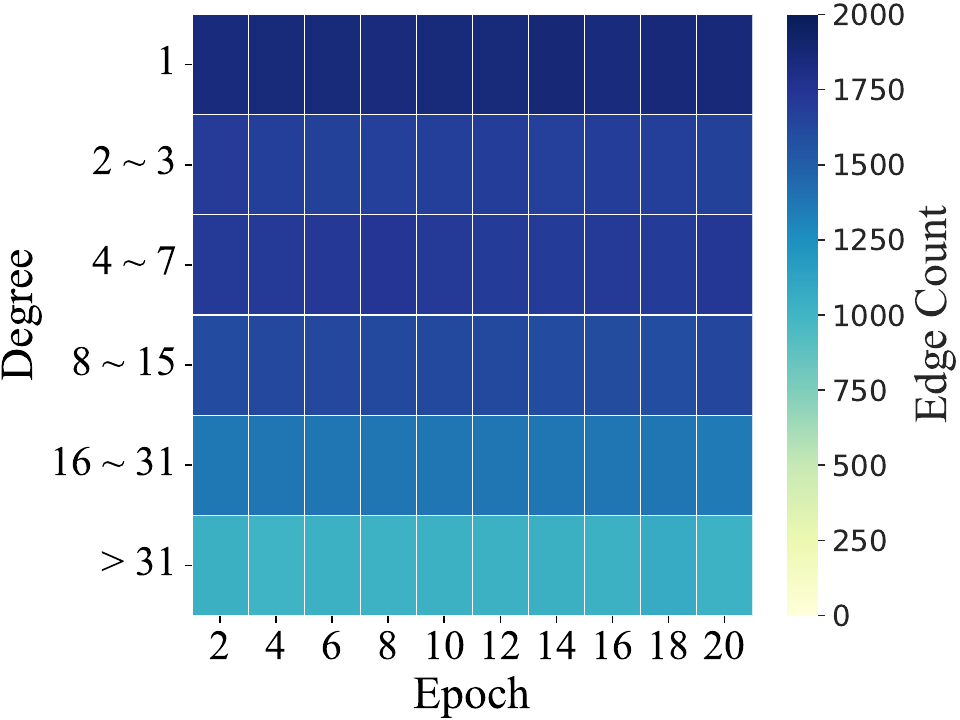}
    \end{minipage}
  }
  \subfigure[GACN]{
    \label{fig:degree_best}
    \begin{minipage}[]{0.475\linewidth}
      \includegraphics[width=1.0\linewidth]{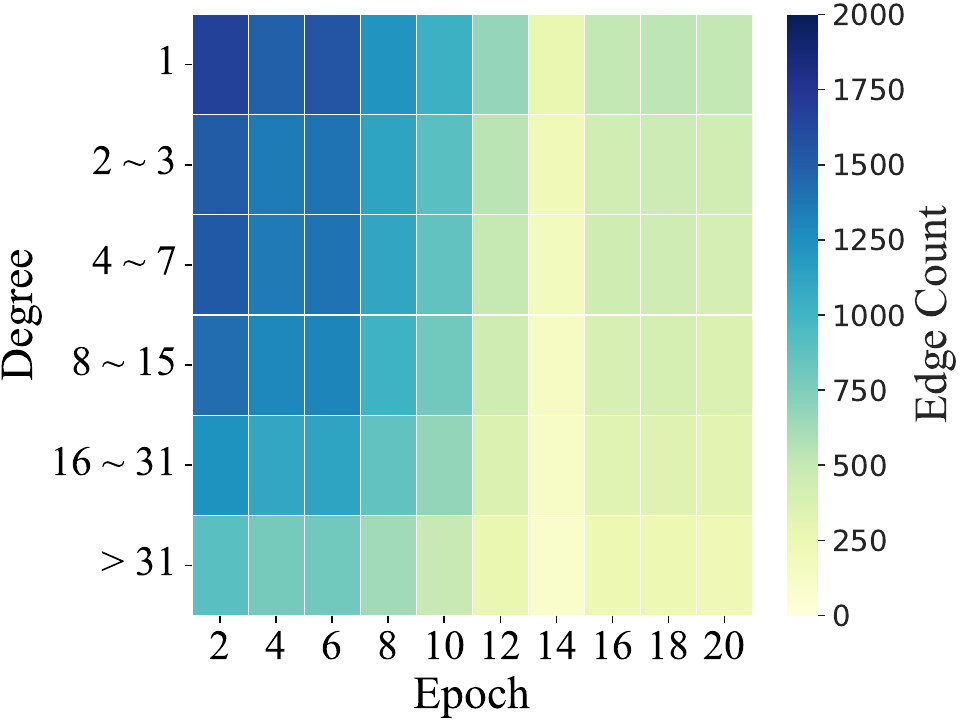}
    \end{minipage}
  }
  \caption{The experimental results of the distribution of new edges.}
  \label{fig:degree}
\end{figure}

\subsection{Quality of Generated Graphs (RQ4)}
\label{sec:case}
In this subsection, we investigate the quality of graphs generated by the view generator w.r.t. \textit{Impact of $\lambda_{cnt}$ and $\lambda_{new}$}, \textit{New Edge Distribution} and \textit{Case Study}.

\subsubsection{Impact of $\lambda_{cnt}$ and $\lambda_{new}$}
In this part, we run the link prediction task on UCI under different settings of $\lambda_{cnt}$ and $\lambda_{new}$, while in each training epoch, we generate ten views and calculate the average amount of edges and the new edges.
As shown in Figure~\ref{fig:new_count}, $\lambda_{cnt}$ contributes to the stability of the edge count, while $\lambda_{new}$ helps to limit the number of new edges.
Specifically, when $\lambda_{cnt} = 1$ and $\lambda_{new} = 0.5$, the number of new edges decreases gradually as the training goes on while the edge count remains stable, which obtains the highest MRR compared to other settings.

\begin{figure*}[ht]
  \centering
  \subfigure[Case 1]{
    \label{fig:case_1}
    \begin{minipage}[]{0.22\linewidth}
      \includegraphics[width=1.0\linewidth]{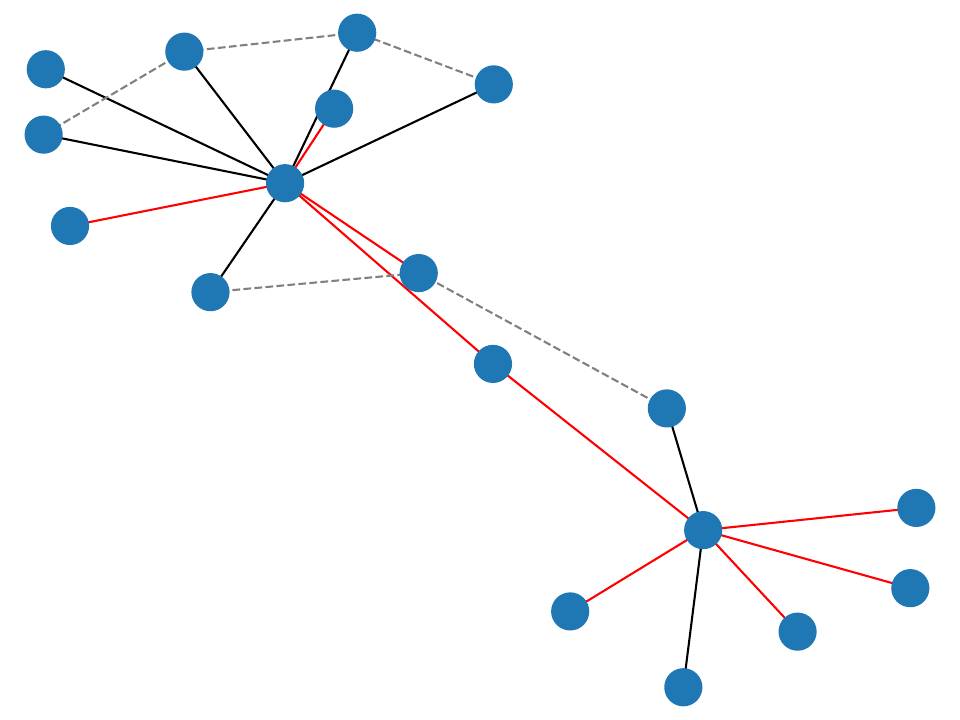}
    \end{minipage}
  }
  \subfigure[Case 2]{
    \label{fig:case_2}
    \begin{minipage}[]{0.22\linewidth}
      \includegraphics[width=1.0\linewidth]{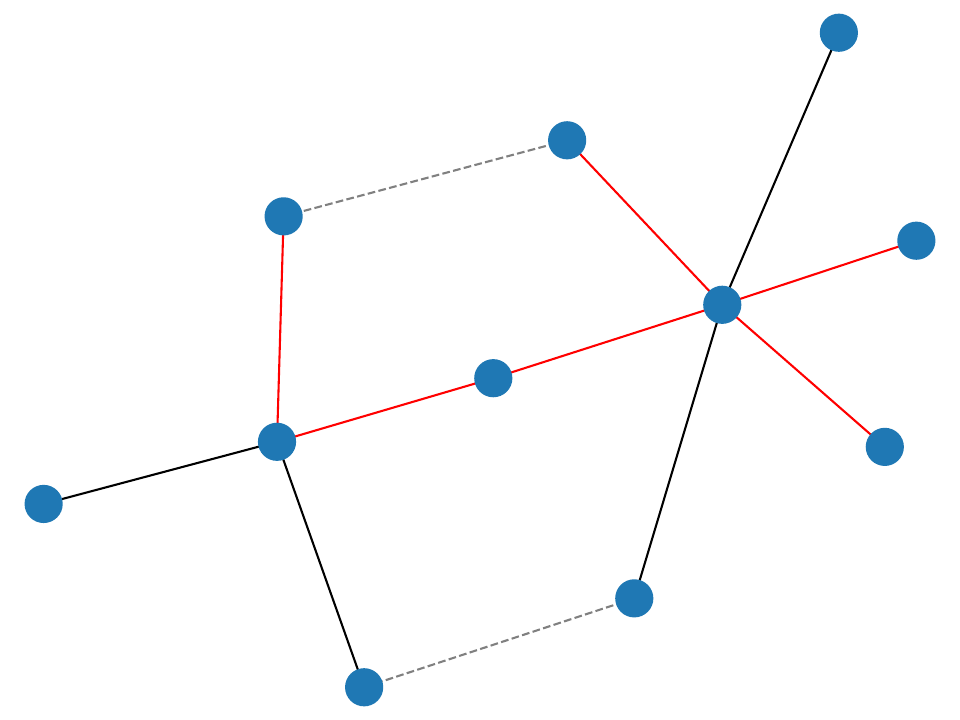}
    \end{minipage}
  }
  \subfigure[Case 3]{
    \label{fig:case_3}
    \begin{minipage}[]{0.22\linewidth}
      \includegraphics[width=1.0\linewidth]{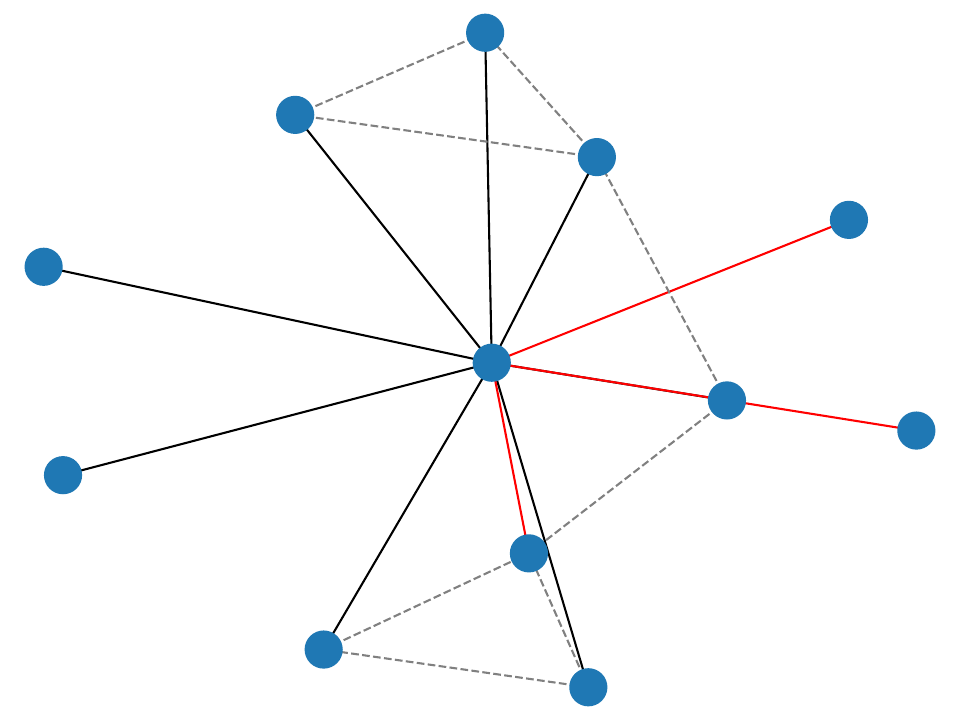}
    \end{minipage}
  }
  \subfigure[Case 4]{
    \label{fig:case_4}
    \begin{minipage}[]{0.22\linewidth}
      \includegraphics[width=1.0\linewidth]{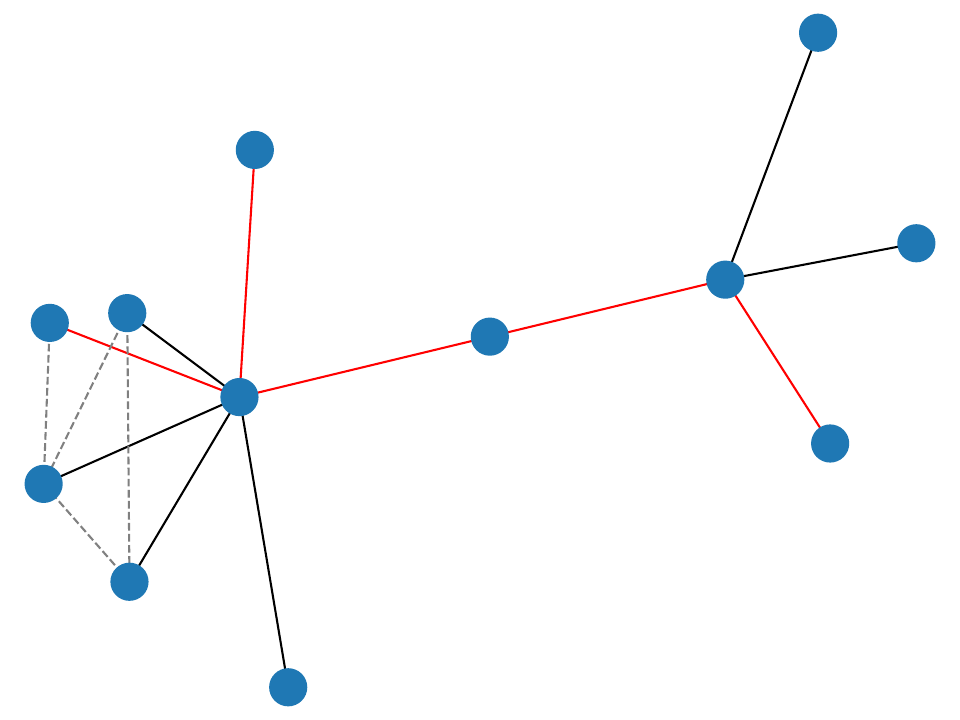}
    \end{minipage}
  }
  \caption{
  Each case is the neighborhood of a sampled node in the generated view.
  Edges in red are generated by GACN. Edges in black are those existing in the original graph and preserved by GACN. 
  Edges in gray and dashed are those existing in the original graph but dropped by GACN.
  It is observed that GACN learns the preferential attachment rule and tends to attach nodes to those with high degree.}
  \label{fig:case_study}
\end{figure*}

\begin{figure*}[ht]
  \centering
  \subfigure[Impact of $s$]{
    \label{fig:ps_s}
    \begin{minipage}[]{0.22\linewidth}
      \includegraphics[width=1.0\linewidth]{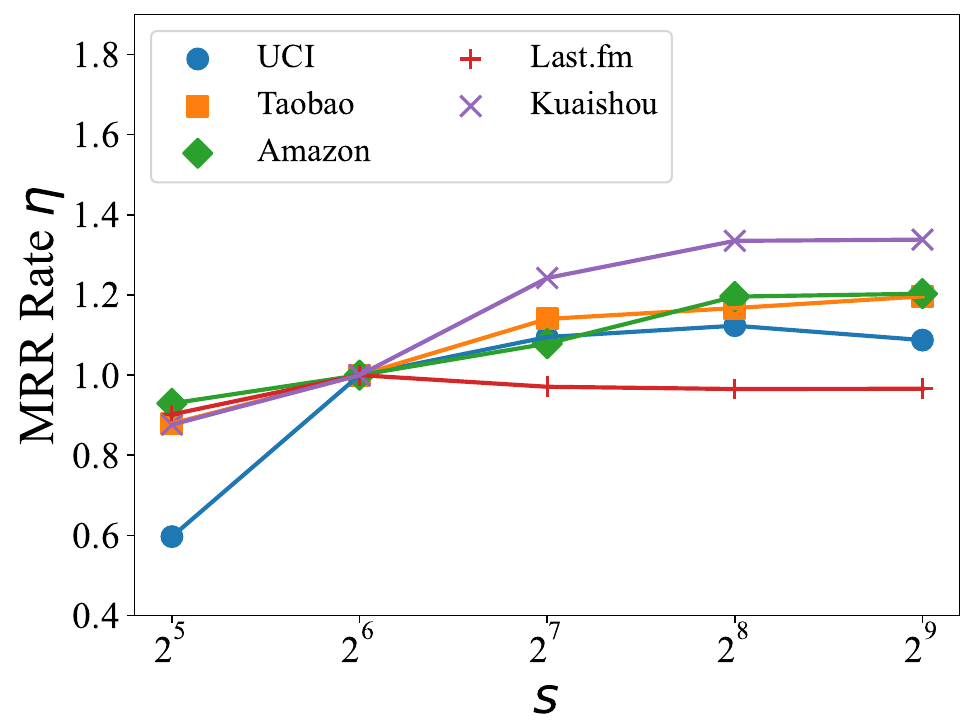}
    \end{minipage}
  }
  \subfigure[Impact of $\tau_g$]{
    \label{fig:ps_tau_g}
    \begin{minipage}[]{0.22\linewidth}
      \includegraphics[width=1.0\linewidth]{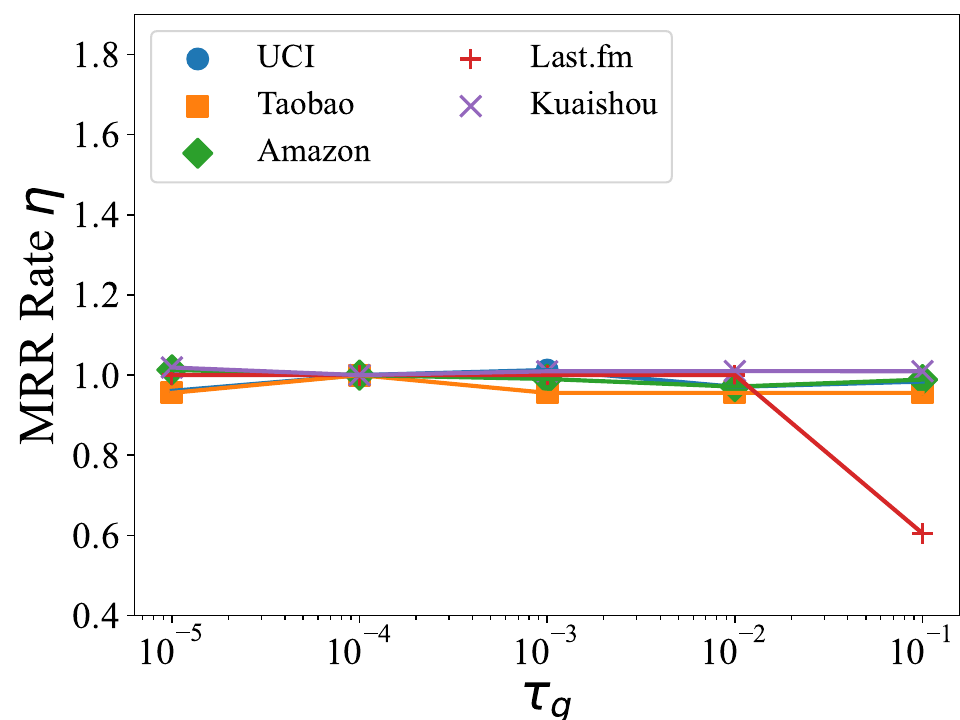}
    \end{minipage}
  }
  \subfigure[Impact of $\lambda_g$]{
    \label{fig:ps_lambda_g}
    \begin{minipage}[]{0.22\linewidth}
      \includegraphics[width=1.0\linewidth]{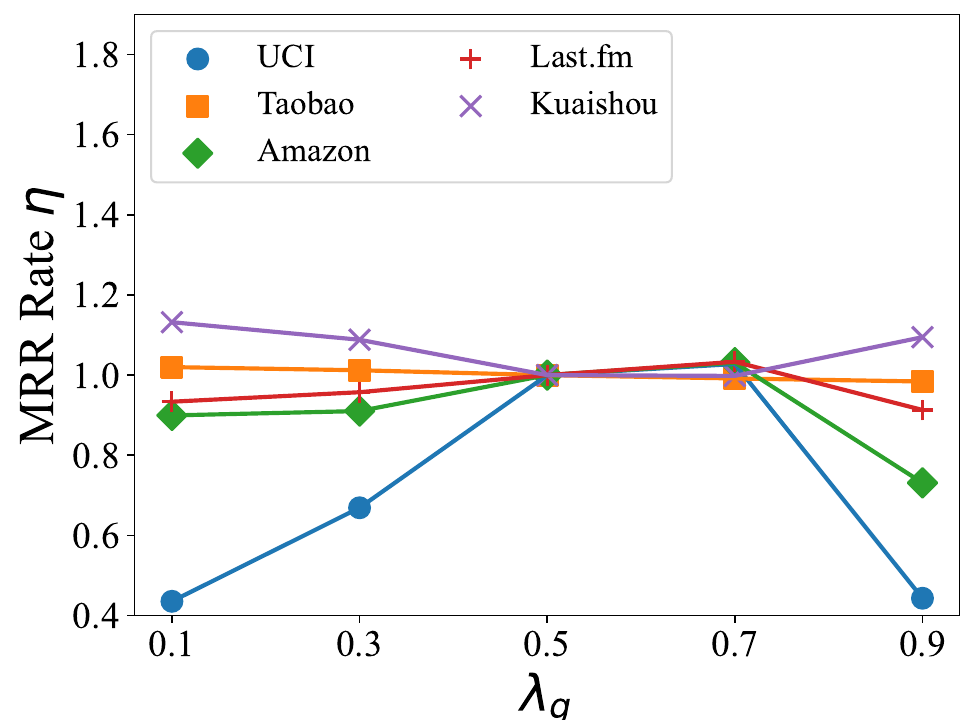}
    \end{minipage}
  }
  \subfigure[Impact of $\tau_f$]{
    \label{fig:ps_tau_f}
    \begin{minipage}[]{0.22\linewidth}
      \includegraphics[width=1.0\linewidth]{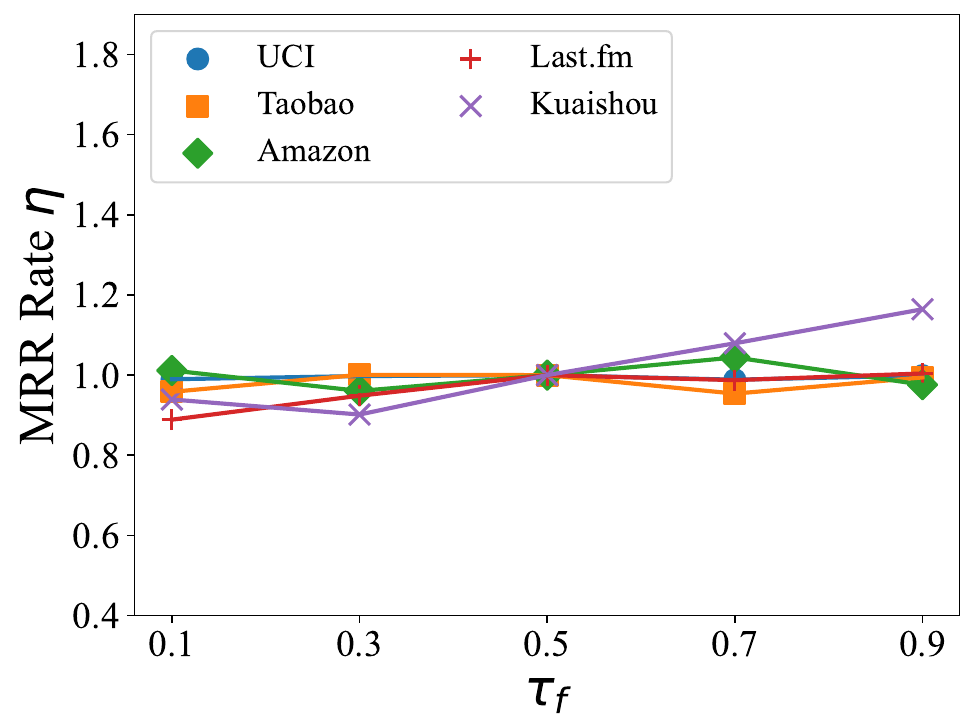}
    \end{minipage}
  }
  \subfigure[Impact of $\lambda_{cnt}$]{
    \label{fig:ps_lambda_cnt}
    \begin{minipage}[]{0.22\linewidth}
      \includegraphics[width=1.0\linewidth]{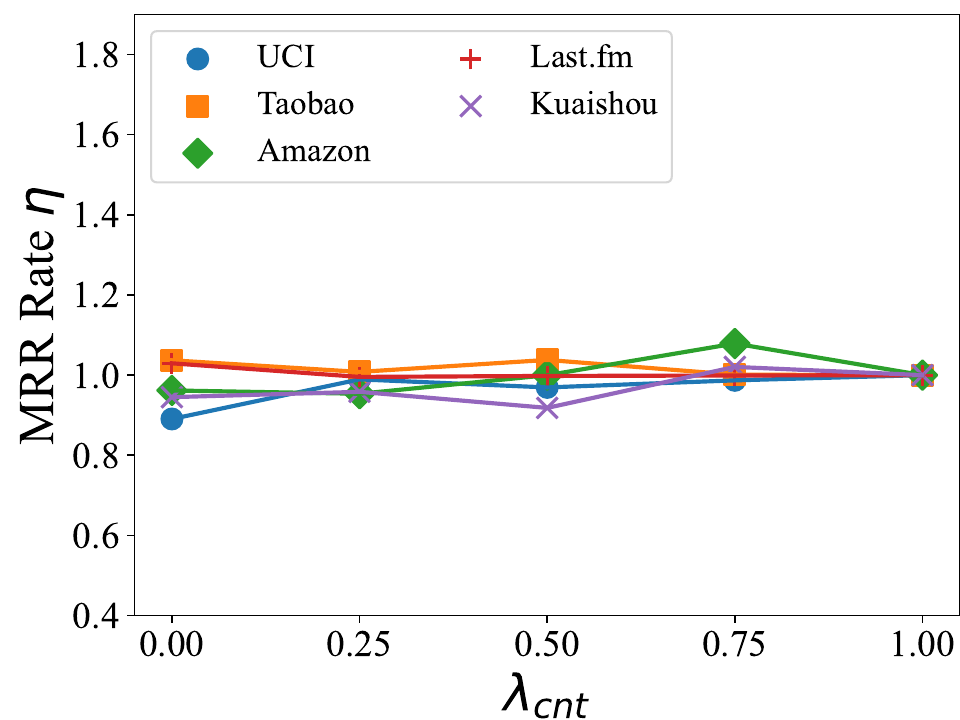}
    \end{minipage}
  }
  \subfigure[Impact of $\lambda_{new}$]{
    \label{fig:ps_lambda_new}
    \begin{minipage}[]{0.22\linewidth}
      \includegraphics[width=1.0\linewidth]{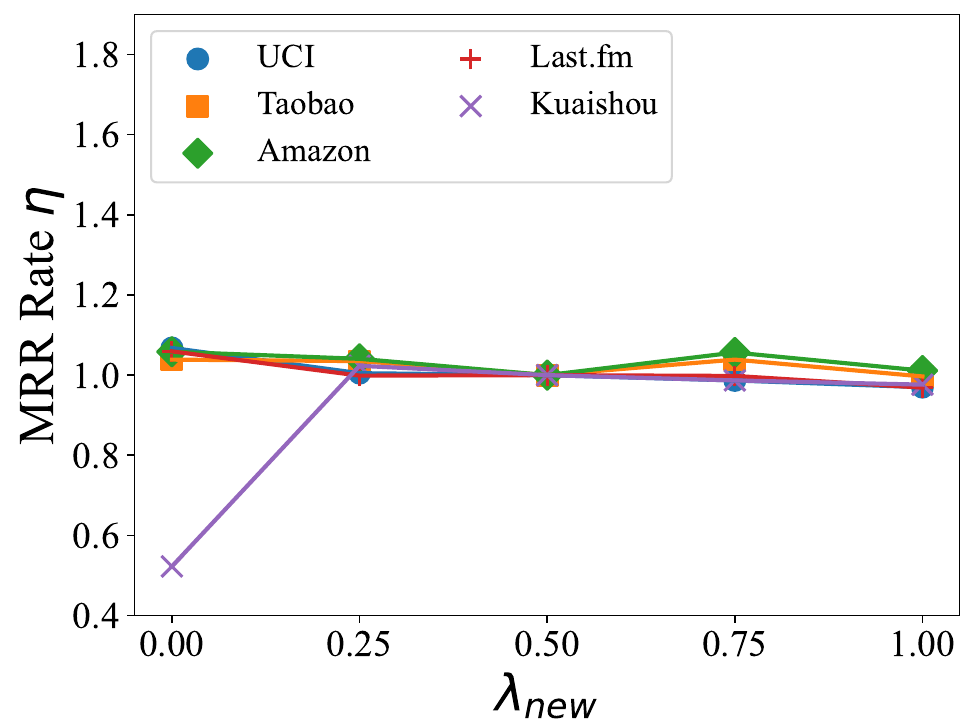}
    \end{minipage}
  }
  \subfigure[Impact of $\gamma$]{
    \label{fig:ps_gamma}
    \begin{minipage}[]{0.22\linewidth}
      \includegraphics[width=1.0\linewidth]{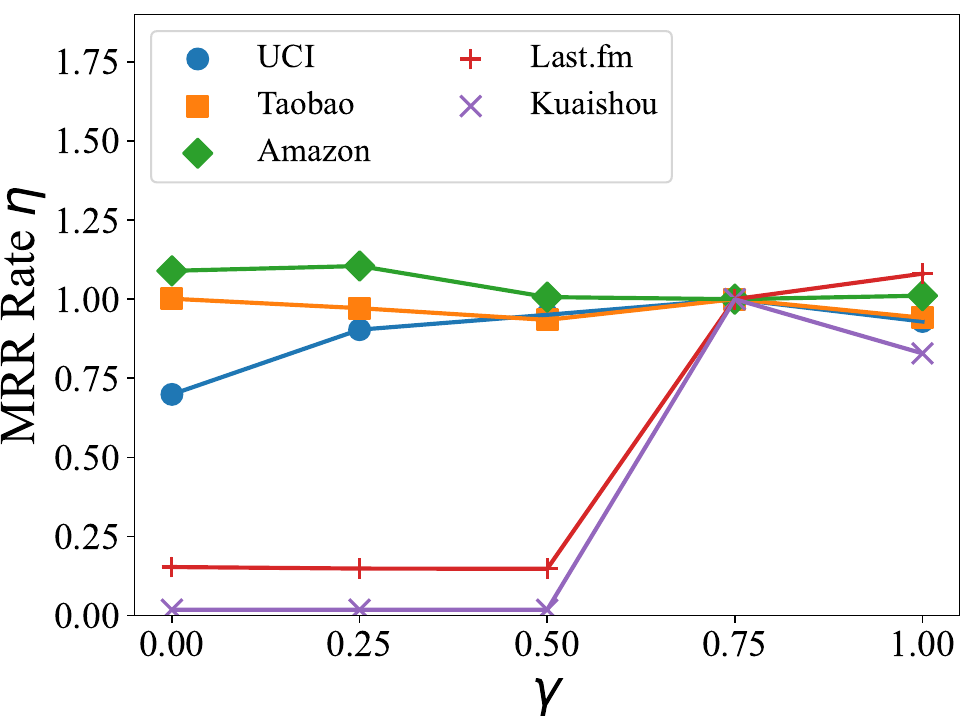}
    \end{minipage}
  }
  \caption{The experimental results of parameter sensitivity.}
  \label{fig:parameter_sensitivity}
\end{figure*}

\subsubsection{New Edge Distribution}
To further investigate the distribution of new edges, we count the number of new edges in groups divided by node degree.
Figure~\ref{fig:degree} illustrates that compared to randomly adding new edges, GACN 1) is able to adjust the number of new edges during training, and 2) generates more edges for high degree nodes (i.e., the color of high degree nodes is more similar to low degree nodes in Figure~\ref{fig:degree_best} compared to that in Figure~\ref{fig:degree_random}), which is in agreement with the preferential attachment rule~\citep{barabasi1999emergence}. 

\subsubsection{Case Study}
For better insight of the views generated by GACN, we randomly sample some nodes in UCI and visualize their neighborhoods within two hops. 
Figure~\ref{fig:case_study} shows that GACN tends to attach nodes to those with high degree and removes other edges, which confirms that GACN indeed learns the preferential attachment rule~\citep{barabasi1999emergence} and is able to generate reasonable alternative views for contrastive learning.


\subsection{Parameter Sensitivity (RQ5)}
In this subsection, we analyze the sensitivity of hyper-parameters in GACN. Specifically, we first examine the impact of the dimension of embedding $s$, the hyper-parameter of the view generator $\tau_g$ and $\lambda_{g}$, and the temperature of the contrastive learning $\tau_f$.
We report the MRR rate $\eta = \frac{\mbox{MRR with current settings}}{\mbox{MRR with default settings}}$
w.r.t. the link prediction task in Figure~\ref{fig:parameter_sensitivity}.

As shown in Figure \ref{fig:ps_s}, the larger dimension of embedding yields the better performance due to the strengthened expression capability of the GACN model.
Figure~\ref{fig:ps_tau_g} shows that GACN is insensitive to $\tau_g$. However, a large $\tau_g$ may result in poor performance.
As shown in Figure~\ref{fig:ps_lambda_g}, GACN is sensitive to $\lambda_g$. Specifically, a small $\lambda_g$ results in sparse views, which are uninformative for contrastive learning, while a large $\lambda_g$ yields dense views, which do harm to robust node representation learning. Generally, setting $\lambda_g$ to $[0.5, 0.7]$ is a good choice.
From Figure~\ref{fig:ps_tau_f}, it is observed that different datasets require different $\tau_f$ for best performance. 
In general, setting $\tau_f$ to $0.5$ yields competitive performances.

We also analyze how the view generator influences GCL, i.e., the sensitivity of $\lambda_{cnt}, \lambda_{new}$ and $\gamma$. It is observed that GACN setting $\lambda_{cnt}$ to $1$ can obtain competitive results (see Figure~\ref{fig:ps_lambda_cnt}), while a small $\lambda_{new}$ is preferred (see Figure~\ref{fig:ps_lambda_new}). However, setting $\lambda_{new}$ to $0$ generates a large amount of unseen edges, and can result in poor performance in some datasets. Thus setting $\lambda_{new}$ to $0.25$ is a better choice. In contrast, different datasets are sensitive to $\gamma$ (see Figure~\ref{fig:ps_gamma}). However, setting $\gamma$ to $0.75$ produces satisfying results.





\section{Conclusion}
\label{sec:conclusion}
In this paper, we incorporated graph GANs with GCL w.r.t. node-level tasks, and presented GACN, a new GNN model that leveraged a graph GAN to generate augmented views for GCL.
Specifically, GACN developed a view generator, a view discriminator and a graph encoder to learn node representations in a self-supervised learning style.
Besides, a novel optimization framework was proposed to train the modules of GACN jointly.
Through comprehensive experiments on seven real-world datasets, we empirically showed the superiority of GACN.
In the future, GACN will be developed to deal with heterogeneous and dynamic graphs.

\begin{acks}
\change{This work is supported in part by the National Natural Science Foundation of China (No.~61872207) and Kuaishou Inc.
Chaokun Wang is the corresponding author.}
\end{acks}


\bibliographystyle{ACM-Reference-Format}
\bibliography{GACoN}


\begin{thebibliography}{54}


\ifx \showCODEN    \undefined \def \showCODEN     #1{\unskip}     \fi
\ifx \showDOI      \undefined \def \showDOI       #1{#1}\fi
\ifx \showISBNx    \undefined \def \showISBNx     #1{\unskip}     \fi
\ifx \showISBNxiii \undefined \def \showISBNxiii  #1{\unskip}     \fi
\ifx \showISSN     \undefined \def \showISSN      #1{\unskip}     \fi
\ifx \showLCCN     \undefined \def \showLCCN      #1{\unskip}     \fi
\ifx \shownote     \undefined \def \shownote      #1{#1}          \fi
\ifx \showarticletitle \undefined \def \showarticletitle #1{#1}   \fi
\ifx \showURL      \undefined \def \showURL       {\relax}        \fi
\providecommand\bibfield[2]{#2}
\providecommand\bibinfo[2]{#2}
\providecommand\natexlab[1]{#1}
\providecommand\showeprint[2][]{arXiv:#2}

\bibitem[kon(2016)]%
        {konect:2016:opsahl-ucsocial}
 \bibinfo{year}{2016}\natexlab{}.
\newblock \bibinfo{title}{UC Irvine messages network dataset -- {KONECT}}.
\newblock
\newblock
\urldef\tempurl%
\url{http://konect.uni-koblenz.de/networks/opsahl-ucsocial}
\showURL{%
\tempurl}


\bibitem[Barab{\'a}si and Albert(1999)]%
        {barabasi1999emergence}
\bibfield{author}{\bibinfo{person}{Albert-L{\'a}szl{\'o} Barab{\'a}si} {and}
  \bibinfo{person}{R{\'e}ka Albert}.} \bibinfo{year}{1999}\natexlab{}.
\newblock \showarticletitle{Emergence of scaling in random networks}.
\newblock \bibinfo{journal}{\emph{science}} \bibinfo{volume}{286},
  \bibinfo{number}{5439} (\bibinfo{year}{1999}), \bibinfo{pages}{509--512}.
\newblock


\bibitem[Becker and Hinton(1992)]%
        {1992Self}
\bibfield{author}{\bibinfo{person}{S. Becker} {and} \bibinfo{person}{G.~E.
  Hinton}.} \bibinfo{year}{1992}\natexlab{}.
\newblock \showarticletitle{Self-organizing neural network that discovers
  surfaces in random-dot stereograms}.
\newblock \bibinfo{journal}{\emph{Nature}} \bibinfo{volume}{355},
  \bibinfo{number}{6356} (\bibinfo{year}{1992}), \bibinfo{pages}{161}.
\newblock


\bibitem[Bojchevski et~al\mbox{.}(2018)]%
        {bojchevski2018netgan}
\bibfield{author}{\bibinfo{person}{Aleksandar Bojchevski},
  \bibinfo{person}{Oleksandr Shchur}, \bibinfo{person}{Daniel Z{\"u}gner},
  {and} \bibinfo{person}{Stephan G{\"u}nnemann}.}
  \bibinfo{year}{2018}\natexlab{}.
\newblock \showarticletitle{Netgan: Generating graphs via random walks}. In
  \bibinfo{booktitle}{\emph{International Conference on Machine Learning}}.
  PMLR, \bibinfo{pages}{610--619}.
\newblock


\bibitem[Cen et~al\mbox{.}(2019)]%
        {cen2019representation}
\bibfield{author}{\bibinfo{person}{Yukuo Cen}, \bibinfo{person}{Xu Zou},
  \bibinfo{person}{Jianwei Zhang}, \bibinfo{person}{Hongxia Yang},
  \bibinfo{person}{Jingren Zhou}, {and} \bibinfo{person}{Jie Tang}.}
  \bibinfo{year}{2019}\natexlab{}.
\newblock \showarticletitle{Representation Learning for Attributed Multiplex
  Heterogeneous Network}. In \bibinfo{booktitle}{\emph{KDD}}.
  \bibinfo{publisher}{ACM}, \bibinfo{pages}{1358--1368}.
\newblock


\bibitem[Chen et~al\mbox{.}(2020)]%
        {chen2020simple}
\bibfield{author}{\bibinfo{person}{Ting Chen}, \bibinfo{person}{Simon
  Kornblith}, \bibinfo{person}{Mohammad Norouzi}, {and}
  \bibinfo{person}{Geoffrey Hinton}.} \bibinfo{year}{2020}\natexlab{}.
\newblock \showarticletitle{A simple framework for contrastive learning of
  visual representations}. In \bibinfo{booktitle}{\emph{International
  conference on machine learning}}. PMLR, \bibinfo{pages}{1597--1607}.
\newblock


\bibitem[Dai et~al\mbox{.}(2018)]%
        {dai2018adversarial}
\bibfield{author}{\bibinfo{person}{Quanyu Dai}, \bibinfo{person}{Qiang Li},
  \bibinfo{person}{Jian Tang}, {and} \bibinfo{person}{Dan Wang}.}
  \bibinfo{year}{2018}\natexlab{}.
\newblock \showarticletitle{Adversarial network embedding}. In
  \bibinfo{booktitle}{\emph{Proceedings of the AAAI Conference on Artificial
  Intelligence}}, Vol.~\bibinfo{volume}{32}.
\newblock


\bibitem[Denton et~al\mbox{.}(2015)]%
        {denton2015deep}
\bibfield{author}{\bibinfo{person}{Emily~L Denton}, \bibinfo{person}{Soumith
  Chintala}, \bibinfo{person}{Rob Fergus}, {et~al\mbox{.}}}
  \bibinfo{year}{2015}\natexlab{}.
\newblock \showarticletitle{Deep generative image models using a laplacian
  pyramid of adversarial networks}.
\newblock \bibinfo{journal}{\emph{Advances in neural information processing
  systems}}  \bibinfo{volume}{28} (\bibinfo{year}{2015}).
\newblock


\bibitem[Gao et~al\mbox{.}(2019)]%
        {gao2019progan}
\bibfield{author}{\bibinfo{person}{Hongchang Gao}, \bibinfo{person}{Jian Pei},
  {and} \bibinfo{person}{Heng Huang}.} \bibinfo{year}{2019}\natexlab{}.
\newblock \showarticletitle{Progan: Network embedding via proximity generative
  adversarial network}. In \bibinfo{booktitle}{\emph{Proceedings of the 25th
  ACM SIGKDD International Conf. on Knowledge Discovery \& Data Mining}}.
  \bibinfo{pages}{1308--1316}.
\newblock


\bibitem[Giles et~al\mbox{.}(1998)]%
        {giles1998citeseer}
\bibfield{author}{\bibinfo{person}{C~Lee Giles}, \bibinfo{person}{Kurt~D
  Bollacker}, {and} \bibinfo{person}{Steve Lawrence}.}
  \bibinfo{year}{1998}\natexlab{}.
\newblock \showarticletitle{CiteSeer: An automatic citation indexing system}.
  In \bibinfo{booktitle}{\emph{Proceedings of the third ACM conference on
  Digital libraries}}. \bibinfo{pages}{89--98}.
\newblock


\bibitem[Goodfellow et~al\mbox{.}(2014)]%
        {goodfellow2014generative}
\bibfield{author}{\bibinfo{person}{Ian Goodfellow}, \bibinfo{person}{Jean
  Pouget-Abadie}, \bibinfo{person}{Mehdi Mirza}, \bibinfo{person}{Bing Xu},
  \bibinfo{person}{David Warde-Farley}, \bibinfo{person}{Sherjil Ozair},
  \bibinfo{person}{Aaron Courville}, {and} \bibinfo{person}{Yoshua Bengio}.}
  \bibinfo{year}{2014}\natexlab{}.
\newblock \showarticletitle{Generative adversarial nets}.
\newblock \bibinfo{journal}{\emph{Advances in neural information processing
  systems}}  \bibinfo{volume}{27} (\bibinfo{year}{2014}).
\newblock


\bibitem[Grover and Leskovec(2016)]%
        {node2vec-kdd2016}
\bibfield{author}{\bibinfo{person}{Aditya Grover} {and} \bibinfo{person}{Jure
  Leskovec}.} \bibinfo{year}{2016}\natexlab{}.
\newblock \showarticletitle{node2vec: Scalable Feature Learning for Networks}.
  In \bibinfo{booktitle}{\emph{KDD}}. ACM.
\newblock


\bibitem[Gu et~al\mbox{.}(2022)]%
        {gu2022hybridgnn}
\bibfield{author}{\bibinfo{person}{Tiankai Gu}, \bibinfo{person}{Chaokun Wang},
  \bibinfo{person}{Cheng Wu}, \bibinfo{person}{Yunkai Lou},
  \bibinfo{person}{Jingcao Xu}, \bibinfo{person}{Changping Wang},
  \bibinfo{person}{Kai Xu}, \bibinfo{person}{Can Ye}, {and}
  \bibinfo{person}{Yang Song}.} \bibinfo{year}{2022}\natexlab{}.
\newblock \showarticletitle{HybridGNN: Learning Hybrid Representation for
  Recommendation in Multiplex Heterogeneous Networks}. In
  \bibinfo{booktitle}{\emph{ICDE}}. IEEE, \bibinfo{pages}{1355--1367}.
\newblock


\bibitem[Hamilton et~al\mbox{.}(2017)]%
        {hamilton2017inductive}
\bibfield{author}{\bibinfo{person}{William~L. Hamilton}, \bibinfo{person}{Rex
  Ying}, {and} \bibinfo{person}{Jure Leskovec}.}
  \bibinfo{year}{2017}\natexlab{}.
\newblock \showarticletitle{Inductive Representation Learning on Large Graphs}.
  In \bibinfo{booktitle}{\emph{NIPS}}.
\newblock


\bibitem[Hassani and Khasahmadi(2020)]%
        {hassani2020contrastive}
\bibfield{author}{\bibinfo{person}{Kaveh Hassani} {and}
  \bibinfo{person}{Amir~Hosein Khasahmadi}.} \bibinfo{year}{2020}\natexlab{}.
\newblock \showarticletitle{Contrastive multi-view representation learning on
  graphs}. In \bibinfo{booktitle}{\emph{ICML}}. PMLR,
  \bibinfo{pages}{4116--4126}.
\newblock


\bibitem[He and McAuley(2016)]%
        {he2016ups}
\bibfield{author}{\bibinfo{person}{Ruining He} {and} \bibinfo{person}{Julian
  McAuley}.} \bibinfo{year}{2016}\natexlab{}.
\newblock \showarticletitle{Ups and downs: Modeling the visual evolution of
  fashion trends with one-class collaborative filtering}. In
  \bibinfo{booktitle}{\emph{The World Wide Web Conference}}.
  \bibinfo{pages}{507--517}.
\newblock


\bibitem[He et~al\mbox{.}(2020)]%
        {he2020lightgcn}
\bibfield{author}{\bibinfo{person}{Xiangnan He}, \bibinfo{person}{Kuan Deng},
  \bibinfo{person}{Xiang Wang}, \bibinfo{person}{Yan Li},
  \bibinfo{person}{Yongdong Zhang}, {and} \bibinfo{person}{Meng Wang}.}
  \bibinfo{year}{2020}\natexlab{}.
\newblock \showarticletitle{Lightgcn: Simplifying and powering graph
  convolution network for recommendation}. In
  \bibinfo{booktitle}{\emph{SIGIR}}. \bibinfo{pages}{639--648}.
\newblock


\bibitem[Henaff(2020)]%
        {henaff2020data}
\bibfield{author}{\bibinfo{person}{Olivier Henaff}.}
  \bibinfo{year}{2020}\natexlab{}.
\newblock \showarticletitle{Data-efficient image recognition with contrastive
  predictive coding}. In \bibinfo{booktitle}{\emph{International Conference on
  Machine Learning}}. PMLR, \bibinfo{pages}{4182--4192}.
\newblock


\bibitem[Hjelm et~al\mbox{.}(2019)]%
        {hjelm2018learning}
\bibfield{author}{\bibinfo{person}{R~Devon Hjelm}, \bibinfo{person}{Alex
  Fedorov}, \bibinfo{person}{Samuel Lavoie-Marchildon}, \bibinfo{person}{Karan
  Grewal}, \bibinfo{person}{Phil Bachman}, \bibinfo{person}{Adam Trischler},
  {and} \bibinfo{person}{Yoshua Bengio}.} \bibinfo{year}{2019}\natexlab{}.
\newblock \showarticletitle{Learning deep representations by mutual information
  estimation and maximization}. In \bibinfo{booktitle}{\emph{ICLR}}.
\newblock


\bibitem[Hou et~al\mbox{.}(2022)]%
        {10.1145/3534678.3539321}
\bibfield{author}{\bibinfo{person}{Zhenyu Hou}, \bibinfo{person}{Xiao Liu},
  \bibinfo{person}{Yukuo Cen}, \bibinfo{person}{Yuxiao Dong},
  \bibinfo{person}{Hongxia Yang}, \bibinfo{person}{Chunjie Wang}, {and}
  \bibinfo{person}{Jie Tang}.} \bibinfo{year}{2022}\natexlab{}.
\newblock \showarticletitle{GraphMAE: Self-Supervised Masked Graph
  Autoencoders}. In \bibinfo{booktitle}{\emph{Proceedings of the 28th ACM
  SIGKDD Conference on Knowledge Discovery and Data Mining}} (Washington DC,
  USA) \emph{(\bibinfo{series}{KDD '22})}. \bibinfo{publisher}{Association for
  Computing Machinery}, \bibinfo{address}{NY, USA}, \bibinfo{pages}{594–604}.
\newblock
\showISBNx{9781450393850}
\urldef\tempurl%
\url{https://doi.org/10.1145/3534678.3539321}
\showDOI{\tempurl}


\bibitem[Hu et~al\mbox{.}(2019)]%
        {hu2019strategies}
\bibfield{author}{\bibinfo{person}{Weihua Hu}, \bibinfo{person}{Bowen Liu},
  \bibinfo{person}{Joseph Gomes}, \bibinfo{person}{Marinka Zitnik},
  \bibinfo{person}{Percy Liang}, \bibinfo{person}{Vijay Pande}, {and}
  \bibinfo{person}{Jure Leskovec}.} \bibinfo{year}{2019}\natexlab{}.
\newblock \showarticletitle{Strategies for pre-training graph neural networks}.
\newblock \bibinfo{journal}{\emph{arXiv preprint arXiv:1905.12265}}
  (\bibinfo{year}{2019}).
\newblock


\bibitem[Jovanovi{\'c} et~al\mbox{.}(2021)]%
        {jovanovic2021towards}
\bibfield{author}{\bibinfo{person}{Nikola Jovanovi{\'c}}, \bibinfo{person}{Zhao
  Meng}, \bibinfo{person}{Lukas Faber}, {and} \bibinfo{person}{Roger
  Wattenhofer}.} \bibinfo{year}{2021}\natexlab{}.
\newblock \showarticletitle{Towards robust graph contrastive learning}.
\newblock \bibinfo{journal}{\emph{arXiv preprint arXiv:2102.13085}}
  (\bibinfo{year}{2021}).
\newblock


\bibitem[Kipf and Welling(2016)]%
        {kipf2016semi}
\bibfield{author}{\bibinfo{person}{Thomas~N Kipf} {and} \bibinfo{person}{Max
  Welling}.} \bibinfo{year}{2016}\natexlab{}.
\newblock \showarticletitle{Semi-supervised classification with graph
  convolutional networks}.
\newblock \bibinfo{journal}{\emph{arXiv preprint arXiv:1609.02907}}
  (\bibinfo{year}{2016}).
\newblock


\bibitem[Lee et~al\mbox{.}(2021)]%
        {lee2021infomax}
\bibfield{author}{\bibinfo{person}{Kwot~Sin Lee}, \bibinfo{person}{Ngoc-Trung
  Tran}, {and} \bibinfo{person}{Ngai-Man Cheung}.}
  \bibinfo{year}{2021}\natexlab{}.
\newblock \showarticletitle{Infomax-gan: Improved adversarial image generation
  via information maximization and contrastive learning}. In
  \bibinfo{booktitle}{\emph{Proceedings of the IEEE/CVF winter conference on
  applications of computer vision}}. \bibinfo{pages}{3942--3952}.
\newblock


\bibitem[Lei et~al\mbox{.}(2019)]%
        {lei2019gcn}
\bibfield{author}{\bibinfo{person}{Kai Lei}, \bibinfo{person}{Meng Qin},
  \bibinfo{person}{Bo Bai}, \bibinfo{person}{Gong Zhang}, {and}
  \bibinfo{person}{Min Yang}.} \bibinfo{year}{2019}\natexlab{}.
\newblock \showarticletitle{GCN-GAN: A non-linear temporal link prediction
  model for weighted dynamic networks}. In \bibinfo{booktitle}{\emph{IEEE
  INFOCOM 2019-IEEE Conference on Computer Communications}}. IEEE,
  \bibinfo{pages}{388--396}.
\newblock


\bibitem[Li et~al\mbox{.}(2017)]%
        {li2017adversarial}
\bibfield{author}{\bibinfo{person}{Jiwei Li}, \bibinfo{person}{Will Monroe},
  \bibinfo{person}{Tianlin Shi}, \bibinfo{person}{S{\'e}bastien Jean},
  \bibinfo{person}{Alan Ritter}, {and} \bibinfo{person}{Dan Jurafsky}.}
  \bibinfo{year}{2017}\natexlab{}.
\newblock \showarticletitle{Adversarial learning for neural dialogue
  generation}.
\newblock \bibinfo{journal}{\emph{arXiv preprint arXiv:1701.06547}}
  (\bibinfo{year}{2017}).
\newblock


\bibitem[Linsker(1988)]%
        {linsker1988self}
\bibfield{author}{\bibinfo{person}{Ralph Linsker}.}
  \bibinfo{year}{1988}\natexlab{}.
\newblock \showarticletitle{Self-organization in a perceptual network}.
\newblock \bibinfo{journal}{\emph{Computer}} \bibinfo{volume}{21},
  \bibinfo{number}{3} (\bibinfo{year}{1988}), \bibinfo{pages}{105--117}.
\newblock


\bibitem[Liu et~al\mbox{.}(2019)]%
        {liu2019learning}
\bibfield{author}{\bibinfo{person}{Weiyi Liu}, \bibinfo{person}{Pin-Yu Chen},
  \bibinfo{person}{Fucai Yu}, \bibinfo{person}{Toyotaro Suzumura}, {and}
  \bibinfo{person}{Guangmin Hu}.} \bibinfo{year}{2019}\natexlab{}.
\newblock \showarticletitle{Learning graph topological features via GAN}.
\newblock \bibinfo{journal}{\emph{IEEE Access}}  \bibinfo{volume}{7}
  (\bibinfo{year}{2019}), \bibinfo{pages}{21834--21843}.
\newblock


\bibitem[McCallum et~al\mbox{.}(2000)]%
        {mccallum2000automating}
\bibfield{author}{\bibinfo{person}{Andrew~Kachites McCallum},
  \bibinfo{person}{Kamal Nigam}, \bibinfo{person}{Jason Rennie}, {and}
  \bibinfo{person}{Kristie Seymore}.} \bibinfo{year}{2000}\natexlab{}.
\newblock \showarticletitle{Automating the construction of internet portals
  with machine learning}.
\newblock \bibinfo{journal}{\emph{Information Retrieval}} \bibinfo{volume}{3},
  \bibinfo{number}{2} (\bibinfo{year}{2000}), \bibinfo{pages}{127--163}.
\newblock


\bibitem[Oord et~al\mbox{.}(2018)]%
        {oord2018representation}
\bibfield{author}{\bibinfo{person}{Aaron van~den Oord}, \bibinfo{person}{Yazhe
  Li}, {and} \bibinfo{person}{Oriol Vinyals}.} \bibinfo{year}{2018}\natexlab{}.
\newblock \showarticletitle{Representation learning with contrastive predictive
  coding}.
\newblock \bibinfo{journal}{\emph{arXiv preprint arXiv:1807.03748}}
  (\bibinfo{year}{2018}).
\newblock


\bibitem[Pan et~al\mbox{.}(2021)]%
        {pan2021videomoco}
\bibfield{author}{\bibinfo{person}{Tian Pan}, \bibinfo{person}{Yibing Song},
  \bibinfo{person}{Tianyu Yang}, \bibinfo{person}{Wenhao Jiang}, {and}
  \bibinfo{person}{Wei Liu}.} \bibinfo{year}{2021}\natexlab{}.
\newblock \showarticletitle{Videomoco: Contrastive video representation
  learning with temporally adversarial examples}. In
  \bibinfo{booktitle}{\emph{Proceedings of the IEEE/CVF Conference on Computer
  Vision and Pattern Recognition}}. \bibinfo{pages}{11205--11214}.
\newblock


\bibitem[Perozzi et~al\mbox{.}(2014)]%
        {perozzi2014deepwalk}
\bibfield{author}{\bibinfo{person}{Bryan Perozzi}, \bibinfo{person}{Rami
  Al-Rfou}, {and} \bibinfo{person}{Steven Skiena}.}
  \bibinfo{year}{2014}\natexlab{}.
\newblock \showarticletitle{Deepwalk: Online learning of social
  representations}. In \bibinfo{booktitle}{\emph{KDD}}.
  \bibinfo{pages}{701--710}.
\newblock


\bibitem[Sun et~al\mbox{.}(2019)]%
        {sun2019megan}
\bibfield{author}{\bibinfo{person}{Yiwei Sun}, \bibinfo{person}{Suhang Wang},
  \bibinfo{person}{Tsung-Yu Hsieh}, \bibinfo{person}{Xianfeng Tang}, {and}
  \bibinfo{person}{Vasant Honavar}.} \bibinfo{year}{2019}\natexlab{}.
\newblock \showarticletitle{Megan: A generative adversarial network for
  multi-view network embedding}.
\newblock \bibinfo{journal}{\emph{arXiv preprint arXiv:1909.01084}}
  (\bibinfo{year}{2019}).
\newblock


\bibitem[Suresh et~al\mbox{.}(2021)]%
        {suresh2021adversarial}
\bibfield{author}{\bibinfo{person}{Susheel Suresh}, \bibinfo{person}{Pan Li},
  \bibinfo{person}{Cong Hao}, {and} \bibinfo{person}{Jennifer Neville}.}
  \bibinfo{year}{2021}\natexlab{}.
\newblock \showarticletitle{Adversarial graph augmentation to improve graph
  contrastive learning}.
\newblock \bibinfo{journal}{\emph{Advances in Neural Information Processing
  Systems}}  \bibinfo{volume}{34} (\bibinfo{year}{2021}).
\newblock


\bibitem[Tang et~al\mbox{.}(2015)]%
        {tang2015line}
\bibfield{author}{\bibinfo{person}{Jian Tang}, \bibinfo{person}{Meng Qu},
  \bibinfo{person}{Mingzhe Wang}, \bibinfo{person}{Ming Zhang},
  \bibinfo{person}{Jun Yan}, {and} \bibinfo{person}{Qiaozhu Mei}.}
  \bibinfo{year}{2015}\natexlab{}.
\newblock \showarticletitle{LINE: Large-scale Information Network Embedding.}.
  In \bibinfo{booktitle}{\emph{WWW}}. ACM.
\newblock


\bibitem[Tavakoli et~al\mbox{.}(2017)]%
        {tavakoli2017learning}
\bibfield{author}{\bibinfo{person}{Sahar Tavakoli}, \bibinfo{person}{Alireza
  Hajibagheri}, {and} \bibinfo{person}{Gita Sukthankar}.}
  \bibinfo{year}{2017}\natexlab{}.
\newblock \showarticletitle{Learning social graph topologies using generative
  adversarial neural networks}. In \bibinfo{booktitle}{\emph{International
  Conference on Social Computing, Behavioral-Cultural Modeling \& Prediction}}.
\newblock


\bibitem[Tian et~al\mbox{.}(2020)]%
        {tian2020contrastive}
\bibfield{author}{\bibinfo{person}{Yonglong Tian}, \bibinfo{person}{Dilip
  Krishnan}, {and} \bibinfo{person}{Phillip Isola}.}
  \bibinfo{year}{2020}\natexlab{}.
\newblock \showarticletitle{Contrastive multiview coding}. In
  \bibinfo{booktitle}{\emph{European conference on computer vision}}. Springer,
  \bibinfo{pages}{776--794}.
\newblock


\bibitem[Van~den Oord et~al\mbox{.}(2018)]%
        {van2018representation}
\bibfield{author}{\bibinfo{person}{Aaron Van~den Oord}, \bibinfo{person}{Yazhe
  Li}, {and} \bibinfo{person}{Oriol Vinyals}.} \bibinfo{year}{2018}\natexlab{}.
\newblock \showarticletitle{Representation learning with contrastive predictive
  coding}.
\newblock \bibinfo{journal}{\emph{arXiv e-prints}} (\bibinfo{year}{2018}),
  \bibinfo{pages}{arXiv--1807}.
\newblock


\bibitem[Velickovic et~al\mbox{.}(2019)]%
        {velickovic2019deep}
\bibfield{author}{\bibinfo{person}{Petar Velickovic}, \bibinfo{person}{William
  Fedus}, \bibinfo{person}{William~L Hamilton}, \bibinfo{person}{Pietro
  Li{\`o}}, \bibinfo{person}{Yoshua Bengio}, {and} \bibinfo{person}{R~Devon
  Hjelm}.} \bibinfo{year}{2019}\natexlab{}.
\newblock \showarticletitle{Deep Graph Infomax.}
\newblock \bibinfo{journal}{\emph{ICLR (Poster)}} \bibinfo{volume}{2},
  \bibinfo{number}{3} (\bibinfo{year}{2019}), \bibinfo{pages}{4}.
\newblock


\bibitem[Wang et~al\mbox{.}(2018)]%
        {wang2018graphgan}
\bibfield{author}{\bibinfo{person}{Hongwei Wang}, \bibinfo{person}{Jia Wang},
  \bibinfo{person}{Jialin Wang}, \bibinfo{person}{Miao Zhao},
  \bibinfo{person}{Weinan Zhang}, \bibinfo{person}{Fuzheng Zhang},
  \bibinfo{person}{Xing Xie}, {and} \bibinfo{person}{Minyi Guo}.}
  \bibinfo{year}{2018}\natexlab{}.
\newblock \showarticletitle{Graphgan: Graph representation learning with
  generative adversarial nets}. In \bibinfo{booktitle}{\emph{Proceedings of the
  AAAI conference on artificial intelligence}}, Vol.~\bibinfo{volume}{32}.
\newblock


\bibitem[Wang et~al\mbox{.}(2017)]%
        {wang2017irgan}
\bibfield{author}{\bibinfo{person}{Jun Wang}, \bibinfo{person}{Lantao Yu},
  \bibinfo{person}{Weinan Zhang}, \bibinfo{person}{Yu Gong},
  \bibinfo{person}{Yinghui Xu}, \bibinfo{person}{Benyou Wang},
  \bibinfo{person}{Peng Zhang}, {and} \bibinfo{person}{Dell Zhang}.}
  \bibinfo{year}{2017}\natexlab{}.
\newblock \showarticletitle{Irgan: A minimax game for unifying generative and
  discriminative information retrieval models}. In
  \bibinfo{booktitle}{\emph{Proceedings of the 40th International ACM SIGIR
  conference on Research and Development in Information Retrieval}}.
  \bibinfo{pages}{515--524}.
\newblock


\bibitem[Wu et~al\mbox{.}(2023)]%
        {wu2023SUPA}
\bibfield{author}{\bibinfo{person}{Cheng Wu}, \bibinfo{person}{Chaokun Wang},
  \bibinfo{person}{Jingcao Xu}, \bibinfo{person}{Ziwei Fang},
  \bibinfo{person}{Tiankai Gu}, \bibinfo{person}{Changping Wang},
  \bibinfo{person}{Yang Song}, \bibinfo{person}{Kai Zheng},
  \bibinfo{person}{Xiaowei Wang}, {and} \bibinfo{person}{Guorui Zhou}.}
  \bibinfo{year}{2023}\natexlab{}.
\newblock \showarticletitle{Instant Representation Learning for Recommendation
  over Large Dynamic Graphs}. In \bibinfo{booktitle}{\emph{2023 IEEE 39th
  International Conference on Data Engineering (ICDE)}}. IEEE,
  \bibinfo{pages}{81--94}.
\newblock


\bibitem[Wu et~al\mbox{.}(2021)]%
        {wu2021self}
\bibfield{author}{\bibinfo{person}{Jiancan Wu}, \bibinfo{person}{Xiang Wang},
  \bibinfo{person}{Fuli Feng}, \bibinfo{person}{Xiangnan He},
  \bibinfo{person}{Liang Chen}, \bibinfo{person}{Jianxun Lian}, {and}
  \bibinfo{person}{Xing Xie}.} \bibinfo{year}{2021}\natexlab{}.
\newblock \showarticletitle{Self-supervised graph learning for recommendation}.
  In \bibinfo{booktitle}{\emph{Proceedings of the 44th international ACM SIGIR
  conference on research and development in information retrieval}}.
  \bibinfo{pages}{726--735}.
\newblock


\bibitem[Yang et~al\mbox{.}(2021)]%
        {yang2021graph}
\bibfield{author}{\bibinfo{person}{Longqi Yang}, \bibinfo{person}{Liangliang
  Zhang}, {and} \bibinfo{person}{Wenjing Yang}.}
  \bibinfo{year}{2021}\natexlab{}.
\newblock \showarticletitle{Graph adversarial self-supervised learning}.
\newblock \bibinfo{journal}{\emph{Advances in Neural Information Processing
  Systems}}  \bibinfo{volume}{34} (\bibinfo{year}{2021}),
  \bibinfo{pages}{14887--14899}.
\newblock


\bibitem[Yang et~al\mbox{.}(2019)]%
        {yang2019advanced}
\bibfield{author}{\bibinfo{person}{Min Yang}, \bibinfo{person}{Junhao Liu},
  \bibinfo{person}{Lei Chen}, \bibinfo{person}{Zhou Zhao},
  \bibinfo{person}{Xiaojun Chen}, {and} \bibinfo{person}{Ying Shen}.}
  \bibinfo{year}{2019}\natexlab{}.
\newblock \showarticletitle{An advanced deep generative framework for temporal
  link prediction in dynamic networks}.
\newblock \bibinfo{journal}{\emph{IEEE transactions on cybernetics}}
  \bibinfo{volume}{50}, \bibinfo{number}{12} (\bibinfo{year}{2019}),
  \bibinfo{pages}{4946--4957}.
\newblock


\bibitem[You et~al\mbox{.}(2021)]%
        {you2021graph}
\bibfield{author}{\bibinfo{person}{Yuning You}, \bibinfo{person}{Tianlong
  Chen}, \bibinfo{person}{Yang Shen}, {and} \bibinfo{person}{Zhangyang Wang}.}
  \bibinfo{year}{2021}\natexlab{}.
\newblock \showarticletitle{Graph contrastive learning automated}. In
  \bibinfo{booktitle}{\emph{International Conference on Machine Learning}}.
  PMLR, \bibinfo{pages}{12121--12132}.
\newblock


\bibitem[You et~al\mbox{.}(2020)]%
        {you2020graph}
\bibfield{author}{\bibinfo{person}{Yuning You}, \bibinfo{person}{Tianlong
  Chen}, \bibinfo{person}{Yongduo Sui}, \bibinfo{person}{Ting Chen},
  \bibinfo{person}{Zhangyang Wang}, {and} \bibinfo{person}{Yang Shen}.}
  \bibinfo{year}{2020}\natexlab{}.
\newblock \showarticletitle{Graph contrastive learning with augmentations}.
\newblock \bibinfo{journal}{\emph{Advances in Neural Information Processing
  Systems}}  \bibinfo{volume}{33} (\bibinfo{year}{2020}),
  \bibinfo{pages}{5812--5823}.
\newblock


\bibitem[Yu et~al\mbox{.}(2017)]%
        {yu2017seqgan}
\bibfield{author}{\bibinfo{person}{Lantao Yu}, \bibinfo{person}{Weinan Zhang},
  \bibinfo{person}{Jun Wang}, {and} \bibinfo{person}{Yong Yu}.}
  \bibinfo{year}{2017}\natexlab{}.
\newblock \showarticletitle{Seqgan: Sequence generative adversarial nets with
  policy gradient}. In \bibinfo{booktitle}{\emph{Proceedings of the AAAI
  conference on artificial intelligence}}, Vol.~\bibinfo{volume}{31}.
\newblock


\bibitem[Yu et~al\mbox{.}(2018)]%
        {yu2018learning}
\bibfield{author}{\bibinfo{person}{Wenchao Yu}, \bibinfo{person}{Cheng Zheng},
  \bibinfo{person}{Wei Cheng}, \bibinfo{person}{Charu~C Aggarwal},
  \bibinfo{person}{Dongjin Song}, \bibinfo{person}{Bo Zong},
  \bibinfo{person}{Haifeng Chen}, {and} \bibinfo{person}{Wei Wang}.}
  \bibinfo{year}{2018}\natexlab{}.
\newblock \showarticletitle{Learning deep network representations with
  adversarially regularized autoencoders}. In
  \bibinfo{booktitle}{\emph{Proceedings of the 24th ACM SIGKDD international
  conf. on knowledge discovery \& data mining}}. \bibinfo{pages}{2663--2671}.
\newblock


\bibitem[Zhang et~al\mbox{.}(2017)]%
        {zhang2017aspect}
\bibfield{author}{\bibinfo{person}{Yuan Zhang}, \bibinfo{person}{Regina
  Barzilay}, {and} \bibinfo{person}{Tommi Jaakkola}.}
  \bibinfo{year}{2017}\natexlab{}.
\newblock \showarticletitle{Aspect-augmented adversarial networks for domain
  adaptation}.
\newblock \bibinfo{journal}{\emph{Transactions of the Association for
  Computational Linguistics}}  \bibinfo{volume}{5} (\bibinfo{year}{2017}),
  \bibinfo{pages}{515--528}.
\newblock


\bibitem[Zhu et~al\mbox{.}(2018)]%
        {zhu2018learning}
\bibfield{author}{\bibinfo{person}{Han Zhu}, \bibinfo{person}{Xiang Li},
  \bibinfo{person}{Pengye Zhang}, \bibinfo{person}{Guozheng Li},
  \bibinfo{person}{Jie He}, \bibinfo{person}{Han Li}, {and}
  \bibinfo{person}{Kun Gai}.} \bibinfo{year}{2018}\natexlab{}.
\newblock \showarticletitle{Learning tree-based deep model for recommender
  systems}. In \bibinfo{booktitle}{\emph{KDD}}. ACM,
  \bibinfo{pages}{1079--1088}.
\newblock


\bibitem[Zhu et~al\mbox{.}(2020)]%
        {zhu2020deep}
\bibfield{author}{\bibinfo{person}{Yanqiao Zhu}, \bibinfo{person}{Yichen Xu},
  \bibinfo{person}{Feng Yu}, \bibinfo{person}{Qiang Liu}, \bibinfo{person}{Shu
  Wu}, {and} \bibinfo{person}{Liang Wang}.} \bibinfo{year}{2020}\natexlab{}.
\newblock \showarticletitle{Deep graph contrastive representation learning}.
\newblock \bibinfo{journal}{\emph{arXiv preprint arXiv:2006.04131}}
  (\bibinfo{year}{2020}).
\newblock


\bibitem[Zhu et~al\mbox{.}(2021)]%
        {zhu2021graph}
\bibfield{author}{\bibinfo{person}{Yanqiao Zhu}, \bibinfo{person}{Yichen Xu},
  \bibinfo{person}{Feng Yu}, \bibinfo{person}{Qiang Liu}, \bibinfo{person}{Shu
  Wu}, {and} \bibinfo{person}{Liang Wang}.} \bibinfo{year}{2021}\natexlab{}.
\newblock \showarticletitle{Graph contrastive learning with adaptive
  augmentation}. In \bibinfo{booktitle}{\emph{Proceedings of the Web Conference
  2021}}. \bibinfo{pages}{2069--2080}.
\newblock


\bibitem[Zitnik et~al\mbox{.}(2018)]%
        {zitnik2018prioritizing}
\bibfield{author}{\bibinfo{person}{Marinka Zitnik}, \bibinfo{person}{Rok
  Sosi{\v{c}}}, {and} \bibinfo{person}{Jure Leskovec}.}
  \bibinfo{year}{2018}\natexlab{}.
\newblock \showarticletitle{Prioritizing network communities}.
\newblock \bibinfo{journal}{\emph{Nature communications}} \bibinfo{volume}{9},
  \bibinfo{number}{1} (\bibinfo{year}{2018}), \bibinfo{pages}{1--9}.
\newblock


\end{thebibliography}

\end{document}